\documentclass[10pt,twocolumn,letterpaper]{article}

\usepackage[pagenumbers]{cvpr} 

\usepackage{graphicx}
\usepackage{booktabs}
\usepackage{multirow}
\usepackage[table]{xcolor}
\usepackage{amssymb} 
\usepackage{pifont}

\definecolor{cvprblue}{rgb}{0.21,0.49,0.74}
\usepackage[pagebackref,breaklinks,colorlinks,allcolors=cvprblue]{hyperref}

\def\confName{CVPR}
\def\confYear{2027}

\title{AGuide: Guiding Air-Ground Feature Adaptation with Flow Matching}

\author{
Xianke Wu$^{1}$\quad
Xinyi Li$^{2}$\quad
Minxue Li$^{1}$\quad
Jian Di$^{3}$\quad
Guiyang Luo$^{1}$\quad
Yonglin Tian$^{4}$\thanks{Corresponding author.} \\[4pt]
$^{1}$School of Computer Science, Beijing University of Posts and Telecommunications \\
$^{2}$School of Computer and Network Engineering, Shanxi Datong University \\
$^{3}$Institute of Advanced Technology, University of Science and Technology of China \\
$^{4}$Institute of Automation, Chinese Academy of Sciences \\[4pt]
{\tt\small sinclairwww@bupt.edu.cn, lixinyi@sxdtdx.edu.cn, 2026110840@bupt.cn,} \\
{\tt\small dijian@mail.ustc.edu.cn, luoguiyang@bupt.edu.cn, yonglin.tian@ia.ac.cn}
}

\begin{document}
\maketitle
\begin{abstract}
Air--ground collaborative perception uses aerial observations to extend vehicle sensing coverage and alleviate occlusion. However, viewpoint differences create a cross-platform representation gap that limits the effective use of aerial information. Existing methods focus on depth estimation, feature interaction, and agent association, with less attention to explicit UAV-to-vehicle feature adaptation. UAV features can be adapted before fusion to bridge this gap, but adapters are typically trained with detection loss alone, which supervises adaptation indirectly through the final predictions. To provide explicit cross-platform supervision, we propose AGuide, a dual-granularity Flow Matching framework that provides explicit cross-platform auxiliary supervision through paired velocity prediction. Object-level Flow Matching captures foreground semantics, while token-level Flow Matching accommodates local cross-view differences. Both branches learn to predict UAV-to-vehicle feature displacements along interpolation paths, providing directional guidance for UAV adaptation alongside detection supervision. We further introduce OPV2V-Air, a camera-based benchmark for air--ground collaborative perception. AGuide improves AP@0.7 by 2.5 percentage points on OPV2V-Air and 4.0 percentage points on V2U4Real, demonstrating more accurate detection in both camera-based simulation and real-world LiDAR-based collaboration.
\end{abstract}    

\begin{figure}[t]
    \centering
    \begin{subfigure}[t]{\columnwidth}
        \centering
        \includegraphics[width=0.55\linewidth]{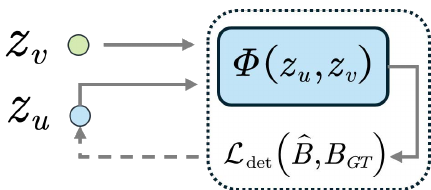}
        \caption{Detection loss supervises UAV representations through the fusion and detection mapping $\Phi$. $\mathbf z_u$ and $\mathbf z_v$ denote UAV and vehicle representations, respectively.}
        \label{fig:motivation_det}
    \end{subfigure}

    \vspace{4pt}
    \begin{subfigure}[t]{\columnwidth}
        \centering
        \includegraphics[width=0.45\linewidth]{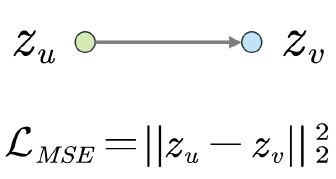}
        \caption{Endpoint MSE directly penalizes the distance between paired UAV and vehicle representations}
        \label{fig:motivation_mse}
    \end{subfigure}

    \vspace{4pt}
    \begin{subfigure}[t]{\columnwidth}
        \centering
        \includegraphics[width=0.65\linewidth]{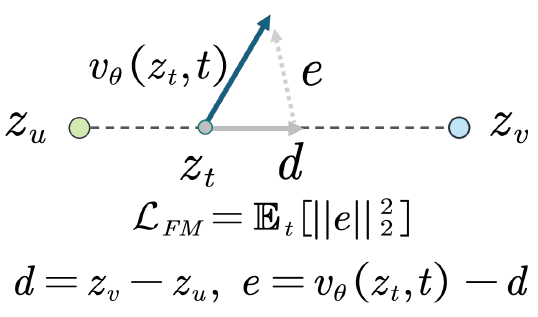}
        \caption{Flow Matching predicts the displacement
$d=z_v-z_u$ at interpolated states
$z_t=(1-t)z_u+tz_v$, where $t\sim\mathcal{U}(0,1)$.
The velocity network $v_\theta$ is parameterized by $\theta$,
and $e$ denotes its prediction error.
}
        \label{fig:motivation_fm}
    \end{subfigure}
    \caption{Different supervision objectives for UAV-to-vehicle feature adaptation. Detection loss provides indirect supervision through task predictions. Endpoint MSE directly constrains paired features to be close. Flow Matching instead supervises velocity predictions using paired feature displacements, providing auxiliary guidance without directly enforcing endpoint agreement.}
    \label{fig:motivation}
\end{figure}

\section{Introduction}

Collaborative perception mitigates the limited sensing range and
occlusions of individual vehicles through multi-agent information
sharing~\cite{xu2022opv2v,wang2020v2vnet}. Owing to their elevated
viewpoints, UAVs provide broad coverage and reduced occlusion,
offering complementary observations of objects that are difficult
to perceive from the ground~\cite{ding2021review,shaharUGVUAVIntegrationAdvancements2025}.
However, differences in sensing viewpoints and observation patterns
create a representation gap between UAVs and ground vehicles,
limiting the effective use of aerial information.
Addressing this gap is therefore important for translating
complementary aerial observations into improvements in vehicle-side
perception.

Existing air--ground collaborative perception methods improve
cross-view collaboration through depth estimation, feature
interaction, distant-object recovery, and cross-agent
association~\cite{10909254,wang2026long,wang2026griffin}.
These approaches improve how aerial and ground observations are
combined, while explicit UAV-to-vehicle feature adaptation before
fusion remains less explored.

Heterogeneous collaborative perception methods address representation differences through agent-specific feature adaptation~\cite{lu2024extensible,gao2025stamp}, providing a foundation for learning compatible representations across sensing platforms. Following this direction, a natural approach is to adapt UAV features before fusion and train the adapter with detection loss. As illustrated in Figure~\ref{fig:motivation}(a), this supervision constrains adaptation indirectly through the final predictions, leaving cross-platform feature correspondence implicit. Endpoint MSE provides explicit feature supervision by directly penalizing the distance between paired UAV and vehicle representations, as shown in Figure~\ref{fig:motivation}(b). However, reducing this distance does not necessarily improve collaborative detection. This motivates explicit cross-platform supervision without directly enforcing feature endpoint agreement.

To this end, we propose AGuide, a dual-granularity Flow Matching framework that guides UAV-to-vehicle feature adaptation through paired velocity prediction. As shown in Figure~\ref{fig:motivation}(c), a velocity network predicts the source-to-reference displacement at sampled interpolation states. This displacement specifies both the direction and magnitude of the target velocity. Prediction errors are backpropagated to the UAV adaptation network, complementing detection loss in learning representations for collaborative detection.

Effective velocity supervision requires appropriate cross-platform reference pairs. We construct these pairs through Reliable ROI Prototype Flow Matching and Deformable Token Flow Matching. At the object level, reliable vehicle proposals define corresponding regions for extracting paired UAV and vehicle prototypes, concentrating supervision on foreground semantics. At the token level, learned sampling offsets and aggregation weights construct reference features from neighboring vehicle tokens. These deformable references accommodate differences in local response locations and spatial extents, complementing the regional semantics captured by object-level prototypes.

To facilitate the exploration of collaborative perception from
vehicle-to-vehicle to air--ground settings, we introduce OPV2V-Air,
a camera-based benchmark built upon OPV2V scenarios.
We extend vehicle-centric scenes with coordinated UAV observations,
synchronized sensor data, and comprehensive 3D annotations.
The benchmark contains 35,313 cooperative timestamps from 68 scenes
and 890,710 UAV 3D bounding-box annotations, providing a large-scale
testbed for studying how aerial information can enhance vehicle-side
perception.

We evaluate AGuide on OPV2V-Air and V2U4Real, covering camera-based
simulation and real-world LiDAR-based collaboration.
On OPV2V-Air, AGuide improves AP@0.7 by 2.5 percentage points over
the strongest compared baseline. On V2U4Real, it improves AP@0.7
by 4.0 percentage points, demonstrating its effectiveness across
the two sensing settings.

Our main contributions are summarized as follows:
\begin{itemize}
\item We propose AGuide, a dual-granularity Flow Matching framework
that provides explicit auxiliary supervision for UAV-to-vehicle
feature adaptation through paired velocity prediction.

\item We introduce Reliable ROI Prototype Flow Matching and
Deformable Token Flow Matching.
The two branches provide complementary supervision for foreground
semantics and local cross-view feature correspondence.

\item We construct OPV2V-Air, a camera-based benchmark for exploring
the extension of vehicle-to-vehicle perception to air--ground
collaboration. Experiments on OPV2V-Air and V2U4Real demonstrate
the effectiveness of AGuide in both camera-based simulation and
real-world LiDAR-based collaboration.
\end{itemize}
\section{Related Work}
\begin{figure*}[t]
    \centering
    \includegraphics[width=\linewidth]{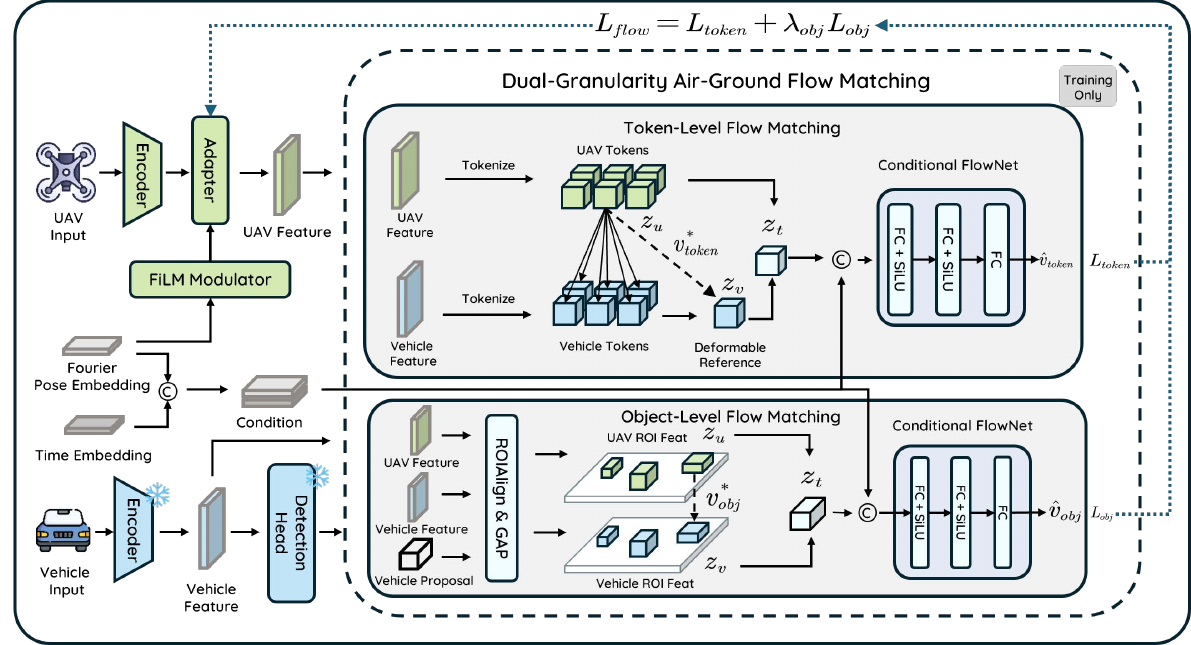}
    \caption{Overview of AGuide. Fourier pose encoding conditions the UAV adapter through FiLM modulation, while a frozen vehicle teacher provides reference features and object proposals. Token-level Flow Matching constructs local references through deformable sampling, and object-level Flow Matching extracts paired ROI representations from selected vehicle proposals. Both branches predict cross-platform feature displacements at sampled interpolation states, providing auxiliary supervision for UAV feature adaptation. The Flow Matching branches are used only during training.}
    \label{fig:method}
\end{figure*}
\subsection{Air--Ground Collaborative Perception}
Several datasets support air--ground collaborative perception. VVSim~\cite{zhu2026vvsim} provides a large-scale aerial--ground dataset and benchmark for cooperative perception, while AGC-Drive~\cite{hou2026agc} and V2U4Real~\cite{Li_2026_CVPR} provide real-world observations for aerial--ground cooperation. Griffin~\cite{wang2026griffin} further supports cooperative detection and tracking under varying communication and sensing conditions.

Air--ground collaborative perception uses UAV observations to extend vehicle-side sensing coverage and alleviate occlusion. AVCPNet~\cite{10909254} improves cross-view interaction and collaborative depth estimation, while Long-SCOPE~\cite{wang2026long} addresses long-range perception through sparse query generation and robust cross-agent association. These methods primarily improve the reconstruction, association, and fusion of cross-view observations. Complementing these efforts, we focus on adapting UAV
representations for vehicle-side fusion and detection.

\subsection{Heterogeneous Collaborative Perception}

 Heterogeneous collaborative perception enables agents with different sensor modalities or model architectures to share information for joint perception. A central challenge is making their representations compatible for feature fusion. MPDA~\cite{10160871}, PnPDA~\cite{luo2024plug}, and PolyInter~\cite{xia2025one} explore feature mappings for heterogeneous agents, while HEAL~\cite{lu2024extensible} and STAMP~\cite{gao2025stamp} enable collaboration through compatible representations. Building on this adaptation paradigm, we investigate explicit
cross-platform supervision to complement detection-driven
learning of UAV representations.

\subsection{Flow Matching for Representation Adaptation}

Flow Matching learns velocity fields along prescribed probability
paths through simulation-free training~\cite{lipman2023flow}.
Beyond generative modeling, FM-KT explores progressive knowledge
transfer between teacher and student representations~\cite{shao2024precise}.
Recent applications also include cross-modal point-cloud
enhancement~\cite{su2026mmwaveflow}, latent feature
denoising~\cite{cao2026c}, and image-space domain
adaptation~\cite{le2026floweo}.
These studies motivate the use of velocity prediction for
cross-platform representation learning.
We employ it as training-only auxiliary supervision over paired
object-level and token-level representations, guiding UAV feature
adaptation for collaborative detection while removing the
velocity networks at inference.
\section{Method}


\subsection{Overview}

We consider air--ground collaborative detection with a vehicle as the ego agent and a UAV as the collaborating agent. Let $\mathbf F^v$ and $\mathbf F^u$ denote their BEV representations. We learn a pose-conditioned UAV adaptation mapping $\mathcal A_\phi$ for collaborative detection:
\begin{equation}
    \widehat{\mathcal B}
    =
    \mathcal D\!\left(
        \mathcal F\!\left(
            \mathbf F^v,\,
            \mathcal A_\phi(\mathbf F^u,\mathbf p^u)
        \right)
    \right),
    \label{eq:collaborative_detection}
\end{equation}
where $\mathbf p^u$ denotes the UAV pose, $\mathcal F$ performs geometrically aligned feature fusion, $\mathcal D$ is the detection head, and $\widehat{\mathcal B}$ is the predicted box set.

As illustrated in Figure~\ref{fig:method}, AGuide complements detection supervision with a training-only Flow Matching objective. We pair adapted UAV representations with vehicle references at two granularities: object-level ROI prototypes and deformable token-level references. Two auxiliary velocity networks predict the displacement between each pair from sampled interpolation states. The prediction errors backpropagate to the UAV adapter, providing cross-platform velocity supervision for learning representations used in collaborative detection.

\subsection{Pose-Conditioned UAV Feature Adaptation}

\subsubsection{Aerial BEV Generation}
High-altitude UAV observations make image-based depth estimation challenging, affecting the quality of the resulting BEV features. Following~\cite{wang2024drones}, we adopt Ground-prior-based BEV Generation (GBG) to incorporate a ground prior into aerial view transformation. The resulting UAV BEV feature $\mathbf F^u$ is then passed to the feature adapter.

\subsubsection{Pose-Conditioned Feature Adaptation}
\label{sec:pose_adaptation}
Following HEAL~\cite{lu2024extensible}, we adopt a ConvNeXt-based adapter to transform UAV BEV features into vehicle-compatible representations. We further condition the adapter on the absolute six-dimensional UAV pose to account for variations in aerial acquisition geometry. The pose is encoded using multi-frequency Fourier features~\cite{tancik2020fourier} and an MLP to capture pose-dependent variations. The resulting encoding modulates the adapter output through feature-wise linear modulation (FiLM):
\begin{equation}
    \mathbf F_a^u
    = \mathcal A_\phi(\mathbf F^u,\mathbf p^u)
    = \operatorname{FiLM}\!\left(
        \mathcal C_\phi(\mathbf F^u),\,
        \mathcal E_p(\mathbf p^u)
    \right),
\end{equation}
where $\mathcal C_\phi$ denotes the ConvNeXt-based adapter, $\mathcal E_p$ is the pose encoder, and $\mathbf p^u$ contains the UAV's absolute position and orientation.


\subsection{Flow-Matching Directional Guidance}

Training the adapter with detection loss supports air--ground collaboration, but supervises UAV adaptation indirectly through the final predictions. We introduce Flow Matching to complement this task-level supervision with a paired velocity prediction objective in feature space.

Flow Matching learns a velocity field by regressing the derivatives of prescribed paths~\cite{lipman2023flow}. For paired endpoints $(\mathbf z_0,\mathbf z_1)\sim\pi$, we use the linear-path objective
\begin{equation}
    \mathcal L_{\mathrm{FM}}
    =
    \mathbb E_{\substack{(\mathbf z_0,\mathbf z_1)\sim\pi\\t\sim\mathcal U(0,1)}}
    \left[
        \left\|
        v_\theta(\mathbf z_t,t)
        -(\mathbf z_1-\mathbf z_0)
        \right\|_2^2
    \right],
    \label{eq:fm_guidance}
\end{equation}
where $\mathbf z_t=(1-t)\mathbf z_0+t\mathbf z_1$ and $v_\theta$ is the velocity network. The target $\mathbf z_1-\mathbf z_0$ is the derivative of the interpolation path, specifying both the direction and magnitude of the source-to-reference displacement.

For UAV adaptation, $\mathbf z_0$ is extracted from the adapted UAV feature $\mathbf F_a^u$, and $\mathbf z_1$ is its paired vehicle reference. These pairs define a cross-platform displacement that the velocity network predicts at sampled interpolation states. We further condition the network on the UAV pose embedding $\mathbf c_p$, yielding $v_\theta(\mathbf z_t,t,\mathbf c_p)$.

The velocity prediction error backpropagates to the UAV adapter, allowing this feature-space auxiliary objective to shape the representations used for collaborative detection. We apply the objective to object-level and token-level pairs through independent velocity networks, with their reference construction described next.

\subsection{Dual-Granularity Reference Construction}

Directional guidance requires pairing UAV representations with vehicle references to define the target velocity $\mathbf z_1-\mathbf z_0$. We construct object-level pairs through ROIAlign and pooling, and token-level pairs through deformable sampling. Each set of pairs supervises an independent velocity network using Eq.~\eqref{eq:fm_guidance}, providing cross-platform velocity supervision at two granularities.

\subsubsection{Object-Level References via ROIAlign}

We select vehicle-teacher proposals whose confidence and ground-truth BEV IoU exceed $\tau_s$ and $\tau_{\mathrm{iou}}$, respectively, forming the proposal set $\mathcal R$. For each proposal $j\in\mathcal R$, we use its box $b_j$ to extract paired object prototypes from UAV and vehicle features aligned in the vehicle ego frame:
\begin{equation}
    \mathbf r_m^j
    =
    \operatorname{GAP}\!\left(
        \operatorname{ROIAlign}(\mathbf F_m,b_j)
    \right),
\end{equation}
where $m\in\{u,v\}$, $\mathbf F_u=\mathbf F_a^u$ and $\mathbf F_v=\mathbf F^v$ denote the aligned adapted UAV and vehicle features, respectively, and $\operatorname{GAP}$ denotes global average pooling.

Teacher proposal filtering selects well-localized vehicle references, while regional pooling aggregates spatial responses into object-level representations. We use $(\mathbf r_u^j,\mathbf r_v^j)$ as the source and reference endpoints in Eq.~\eqref{eq:fm_guidance}, supervising velocity prediction at the scale of an object region.


\begin{table*}[t]
\centering
\small
\setlength{\tabcolsep}{6pt}
\renewcommand{\arraystretch}{1.12}
\caption{Detection performance on OPV2V-Air and V2U4Real. $\ddagger$ denotes our reimplementation.}
\label{tab:overall_results}
\begin{tabular}{llcccccc}
\toprule
\multirow{2}{*}{Method} & \multirow{2}{*}{Publication} & \multicolumn{3}{c}{OPV2V-Air (Camera)} & \multicolumn{3}{c}{V2U4Real (LiDAR)} \\
\cmidrule(lr){3-5}\cmidrule(lr){6-8}
& & AP@0.3 & AP@0.5 & AP@0.7 & AP@0.3 & AP@0.5 & AP@0.7 \\
\midrule
No Fusion (Vehicle) & -- & 57.0 & 43.1 & 23.3 & 47.9 & 47.1 & 30.3 \\
No Fusion (UAV) & -- & 87.5 & 82.5 & 66.3 & 33.2 & 31.3 & 17.1 \\
Late Fusion & -- & 78.0 & 74.4 & 61.2 & 45.2 & 43.2 & 21.7 \\
\midrule
CoBEVT\cite{pmlr-v205-xu23a} & CoRL'22 & 81.2 & 77.0 & 59.4 & 44.3 & 41.8 & 22.2 \\
Where2comm\cite{hu2022where2comm} & NeurIPS'22 & 85.1 & 79.0 & 58.8 & 57.1 & 53.9 & 29.9 \\
V2X-ViT\cite{xu2022v2x} & ECCV'22 & 82.7 & 78.4 & 62.2 & 42.9 & 40.9 & 22.0 \\
CoAlign\cite{10160546} & ICRA'23 & \textbf{87.8} & \underline{83.9} & \underline{72.4} & \underline{58.4} & \underline{56.6} & \underline{36.6} \\
HEAL & ICLR'24 & 85.5 & 83.2 & 67.8 & 47.1 & 46.7 & 32.9 \\
AVCPNet$^\ddagger$ & TGRS'25 & 86.0 & 83.2 & 70.0 & 55.9 & 55.0 & 35.4 \\
\textbf{AGuide (Ours)} & -- & \underline{87.6} & \textbf{85.1} & \textbf{74.9} & \textbf{62.9} & \textbf{61.7} & \textbf{40.6} \\
\bottomrule
\end{tabular}
\end{table*}


\subsubsection{Token-Level References via Deformable Sampling}

Object-level pooling summarizes regional semantics, but local responses may differ in position and extent across viewpoints. We therefore construct vehicle references through learnable local sampling. After tokenizing both aligned feature maps, we predict $K$ sampling offsets and aggregation weights for each foreground UAV token. Its vehicle reference is
\begin{equation}
    \widetilde{\mathbf z}_v^{\,i}
    =
    \sum_{k=1}^{K}
    \alpha_{ik}\,
    \operatorname{Sample}\!\left(
        \mathbf Z_v,\,
        \mathbf q_i+\Delta\mathbf q_{ik}
    \right),
\end{equation}
where $\mathbf Z_v$ is the vehicle token map, $\mathbf q_i$ is the UAV token location, and $\Delta\mathbf q_{ik}$ is a bounded local sampling offset. The weights $\alpha_{ik}$ are normalized by softmax, and $\operatorname{Sample}$ denotes bilinear sampling.

The learned offsets and weights allow each foreground UAV token to aggregate nearby vehicle responses, reducing dependence on exact same-location correspondence. We use the UAV token $\mathbf z_u^i$ and its aggregated vehicle reference $\widetilde{\mathbf z}_v^{\,i}$ as the paired endpoints in Eq.~\eqref{eq:fm_guidance}, providing local velocity supervision.

Deformable sampling is applied only to foreground tokens covered by ground-truth boxes, while background tokens retain same-location vehicle references. Both participate in Flow Matching supervision. We assign decreasing weights to foreground covered by reliable vehicle-teacher proposals, remaining foreground, and background:
\begin{equation}
    w_{\mathrm{veh}}>w_{\mathrm{fg}}>w_{\mathrm{bg}}.
\end{equation}
This weighting prioritizes foreground regions with reliable vehicle references while retaining supervision over the remaining feature map.


\subsection{Conditional Flow Networks}
\label{sec:conditional_flow}

We use independently parameterized conditional MLPs for token-level and object-level velocity prediction:
\begin{equation}
    \widehat{\mathbf v}_b
    =
    \operatorname{MLP}_{\theta_b}\!\left(
        \left[\mathbf z_t^b \,\Vert\, \mathcal E_t(t)
        \,\Vert\, \mathcal E_p^b(\mathbf p^u)\right]
    \right),
\end{equation}
where $b\in\{\mathrm{token},\mathrm{obj}\}$ indexes the two branches, $\mathbf z_t^b$ is the interpolated representation, $\Vert$ denotes concatenation, and $\theta_b$ denotes the branch-specific learnable parameters. We use sinusoidal timestep encoding $\mathcal E_t$ and the Fourier-based pose encoding $\mathcal E_p^b$ described in Section~\ref{sec:pose_adaptation}. The conditioning embeddings are broadcast to the corresponding tokens or ROI prototypes.

Both MLPs use three fully connected layers with SiLU activations between them. Supervised by paired source-to-reference displacements, they provide auxiliary gradients for UAV feature adaptation during training and are removed at inference.


\subsection{Training Objective}

We apply Eq.~\eqref{eq:fm_guidance} to the object-level
and token-level reference pairs. The resulting losses are
aggregated over selected ROIs and weighted tokens to obtain
$\mathcal L_{\mathrm{obj}}$ and $\mathcal L_{\mathrm{token}}$,
respectively. The overall training objective is
\begin{equation}
    \mathcal L
    =
    \mathcal L_{\mathrm{det}}
    +
    \lambda_{\mathrm{FM}}
    \left(
        \mathcal L_{\mathrm{token}}
        +
        \lambda_{\mathrm{obj}}\mathcal L_{\mathrm{obj}}
    \right),
\end{equation}
where $\mathcal L_{\mathrm{det}}$ denotes the detection objective,
$\lambda_{\mathrm{FM}}$ controls the strength of directional
supervision, and $\lambda_{\mathrm{obj}}$ balances the
object-level and token-level branches.

\section{Experiments}
\subsection{Experimental Setup}
\subsubsection{Datasets}
\paragraph{OPV2V-Air.}
We construct OPV2V-Air, a camera-based simulation benchmark
that extends OPV2V driving scenarios with synchronized UAV
observations. We evaluate on its test split, with further
dataset details provided in the Appendix.

\paragraph{V2U4Real.}
V2U4Real\cite{Li_2026_CVPR} is a real-world collaborative perception dataset
with paired vehicle and UAV observations. We train on its publicly available
LiDAR training split and evaluate on the validation split.
This setting examines the applicability of AGuide beyond
camera-based simulation to real-world LiDAR collaboration.

\subsubsection{Evaluation Protocol}

We report BEV AP at IoU thresholds of 0.3, 0.5, and 0.7
on both datasets. The main comparisons use one ego vehicle
and one UAV at inference. Training configurations and
evaluation details are provided in the Appendix.

\subsubsection{Implementation Details}

We implement AGuide on top of HEAL~\cite{lu2024extensible} and follow a three-stage training procedure. We first pretrain the vehicle detector to establish the reference representation space. We then freeze the vehicle teacher and train the UAV adaptation network with detection and auxiliary supervision. Finally, we fine-tune the collaborative model with detection loss. On OPV2V-Air, the latter two stages use 20-epoch schedules with a batch size of four per GPU. We use AdamW with a weight decay of $0.01$, an initial learning rate of $1\times10^{-4}$, and cosine decay. We set $\lambda_{\mathrm{FM}}=0.03$ and $\lambda_{\mathrm{obj}}=1.2$. Vehicle-teacher proposals are selected using a confidence threshold of $0.3$ and a ground-truth BEV IoU threshold of $0.5$, retaining at most 64 proposals per sample. Object prototypes are extracted through $3\times3$ ROIAlign followed by average pooling, while token-level references use nine sampling points per foreground token. Experiments run on NVIDIA GeForce RTX 4090 GPUs. Further details are provided in the Appendix.


\subsection{Quantitative Results}
\subsubsection{Comparison with Existing Methods}

Table~\ref{tab:overall_results} compares AGuide with No Fusion on each platform and Late Fusion as basic baselines, alongside established vehicle--vehicle collaborative perception methods, including CoBEVT, Where2comm, V2X-ViT, and CoAlign. We also include HEAL, a heterogeneous collaborative perception method, and AVCPNet, an air--ground collaborative perception method. Since AVCPNet's official code is unavailable, we reimplement it following the published design, with implementation details provided in the Appendix.

On OPV2V-Air, AGuide achieves comparable AP@0.3 to CoAlign while
attaining the highest AP@0.5 and AP@0.7, with a 2.5-percentage-point
gain in AP@0.7. On V2U4Real, AGuide leads across all three IoU
thresholds, surpassing CoAlign by 4.0 percentage points in AP@0.7.
These real-world LiDAR results complement the camera-based
simulation findings and support the applicability of our framework
across sensing modalities and acquisition settings.

\begin{table}[t]
\centering
\small
\setlength{\tabcolsep}{4pt}
\renewcommand{\arraystretch}{1.08}
\caption{Incremental component ablation on OPV2V-Air. }
\begin{tabular}{@{}lccc@{}}
\toprule
Configuration & AP@0.3 & AP@0.5 & AP@0.7 \\
\midrule
Baseline & 85.5 & 83.2 & 67.8 \\
+ GBG & 86.5 & 83.7 & 69.6 \\
+ Fourier Pose Encoding & 86.9 & 84.3 & 71.8 \\
+ Token-Level Flow Matching & 87.3 & 84.7 & 72.1 \\
+ Object-Level Flow Matching & \textbf{87.6} & \textbf{85.1} & \textbf{74.9} \\
\bottomrule
\end{tabular}
\label{tab:incremental_ablation}
\end{table}

\begin{table}[t]
\centering
\small
\setlength{\tabcolsep}{4pt}
\renewcommand{\arraystretch}{1.08}
\caption{Comparison of auxiliary supervision strategies on
OPV2V-Air. All variants retain the detection loss.
$\mathrm{MMD}^2$ measures the distribution discrepancy between
UAV and vehicle ROI features using a shared RBF kernel bandwidth.}
\begin{tabular}{@{}lccc@{}}
\toprule
Auxiliary Supervision & AP@0.5 & AP@0.7 & $\mathrm{MMD}^2$ \\
\midrule
Detection Loss Only & 84.3 & 71.8 & 0.36 \\
Endpoint MSE & 82.0 & 68.9 & 0.31 \\
Detached Flow Matching & 84.2 & 71.6 & 0.42 \\
Flow Matching & \textbf{85.1} & \textbf{74.9} & 0.39 \\
\bottomrule
\end{tabular}
\label{tab:supervision_analysis}
\end{table}


\begin{table}[t]
\centering
\small
\setlength{\tabcolsep}{4pt}
\renewcommand{\arraystretch}{1.15}
\caption{Ablation of reference construction for Flow Matching supervision on OPV2V-Air. Combining object-level and deformable token-level references achieves the best performance across all IoU thresholds.}
\begin{tabular}{@{}lccc@{}}
\toprule
Reference Configuration & AP@0.3 & AP@0.5 & AP@0.7 \\
\midrule
Map-Level (Dense BEV) & 84.3 & 81.6 & 65.2 \\
Object-Level (ROIAlign) & 87.0 & 84.6 & 73.7 \\
Token-Level (Same-Cell) & 86.9 & 84.0 & 68.8 \\
Token-Level (Deformable) & 87.3 & 84.7 & 72.1 \\
\midrule
Object + Token (Same-Cell) & 87.3 & 85.0 & 73.4 \\
Object + Token (Deformable) & \textbf{87.6} & \textbf{85.1} & \textbf{74.9} \\
\bottomrule
\end{tabular}
\label{tab:reference_ablation}
\end{table}


\begin{table}[t]
\centering
\small
\setlength{\tabcolsep}{5pt}
\renewcommand{\arraystretch}{1.08}
\caption{Ablation of Flow Matching time settings on OPV2V-Air. Fixed and randomly sampled interpolation times both improve detection over the baseline without Flow Matching, with random sampling achieving the highest AP@0.7.}
\begin{tabular}{@{}lccc@{}}
\toprule
Setting & AP@0.3 & AP@0.5 & AP@0.7 \\
\midrule
Without Flow Matching & 86.9 & 84.3 & 71.8 \\
Fixed $t=0$ & 87.1 & 84.4 & 73.8 \\
Fixed $t=0.5$ & \textbf{87.7} & \textbf{85.3} & 74.5 \\
Random $t\sim\mathcal U(0,1)$ & 87.6 & 85.1 & \textbf{74.9} \\
\bottomrule
\end{tabular}
\label{tab:fm_time}
\end{table}

\subsubsection{Ablation Study}

\paragraph{Incremental Component Ablation.}
Table~\ref{tab:incremental_ablation} evaluates the progressive addition of each component. Adding GBG and Fourier pose conditioning raises AP@0.7 from 67.8 to 71.8. Token-level Flow Matching with deformable references further improves it to 72.1, while adding object-level Flow Matching yields 74.9. These results support combining geometry-conditioned adaptation with Flow Matching supervision at both token and object levels.

\paragraph{Ablation of Supervision Strategies.}
Table~\ref{tab:supervision_analysis} compares detection-only
training with three auxiliary supervision strategies, all built
on a baseline incorporating GBG and Fourier pose encoding.
Endpoint MSE directly regresses UAV representations toward paired
vehicle references. Detached Flow Matching trains the auxiliary velocity
networks while blocking Flow Matching gradients to the deployed representation
modules. Flow Matching allows these gradients to guide UAV feature
adaptation.

Endpoint MSE yields a lower $\mathrm{MMD}^2$ but worse detection
performance than detection-only training, indicating that reducing
global feature distribution discrepancy does not necessarily
improve collaborative detection. In contrast, Flow Matching achieves
the highest detection AP without minimizing $\mathrm{MMD}^2$.
Its advantage over Detached Flow Matching is consistent with the value of
allowing the auxiliary supervision to shape UAV representations,
rather than merely fitting the velocity networks.

\paragraph{Ablation of Reference Construction.}
Table~\ref{tab:reference_ablation} compares different reference construction strategies. Map-level supervision pairs dense UAV and vehicle BEV features, while object-level supervision pairs pooled ROI features from reliable vehicle-teacher proposals. At the token level, same-cell references use vehicle tokens at corresponding grid locations, whereas deformable references aggregate nearby vehicle features through learnable sampling. We also combine object-level references with each token-level variant.

Among the individual strategies, object-level references achieved the highest AP@0.7, and deformable token references outperformed same-cell references at all evaluated IoU thresholds. Combining object-level and deformable token references achieved the highest AP across all evaluated IoU thresholds, suggesting that regional object summaries and flexible local correspondences provide complementary supervision.



\paragraph{Analysis of Flow Matching Time Settings.}
Table~\ref{tab:fm_time} compares fixed and randomly sampled interpolation times with the baseline without Flow Matching. Given $\mathbf z_t=(1-t)\mathbf z_u+t\mathbf z_v$, all three settings predict the same displacement $\mathbf z_v-\mathbf z_u$ but use different input states. At $t=0$, the network predicts this displacement from UAV features and pose conditioning, while $t=0.5$ provides a fixed mixture of UAV and vehicle features. Both improve AP across all three IoU thresholds over the baseline, showing that paired velocity prediction can benefit adaptation even at a single path position. Random sampling exposes the network to varying mixtures along the path and achieves the highest AP@0.7, suggesting that supervision across multiple interpolation states further benefits strict-IoU detection.

\begin{figure}[t]
    \centering
    \includegraphics[width=0.7\columnwidth]{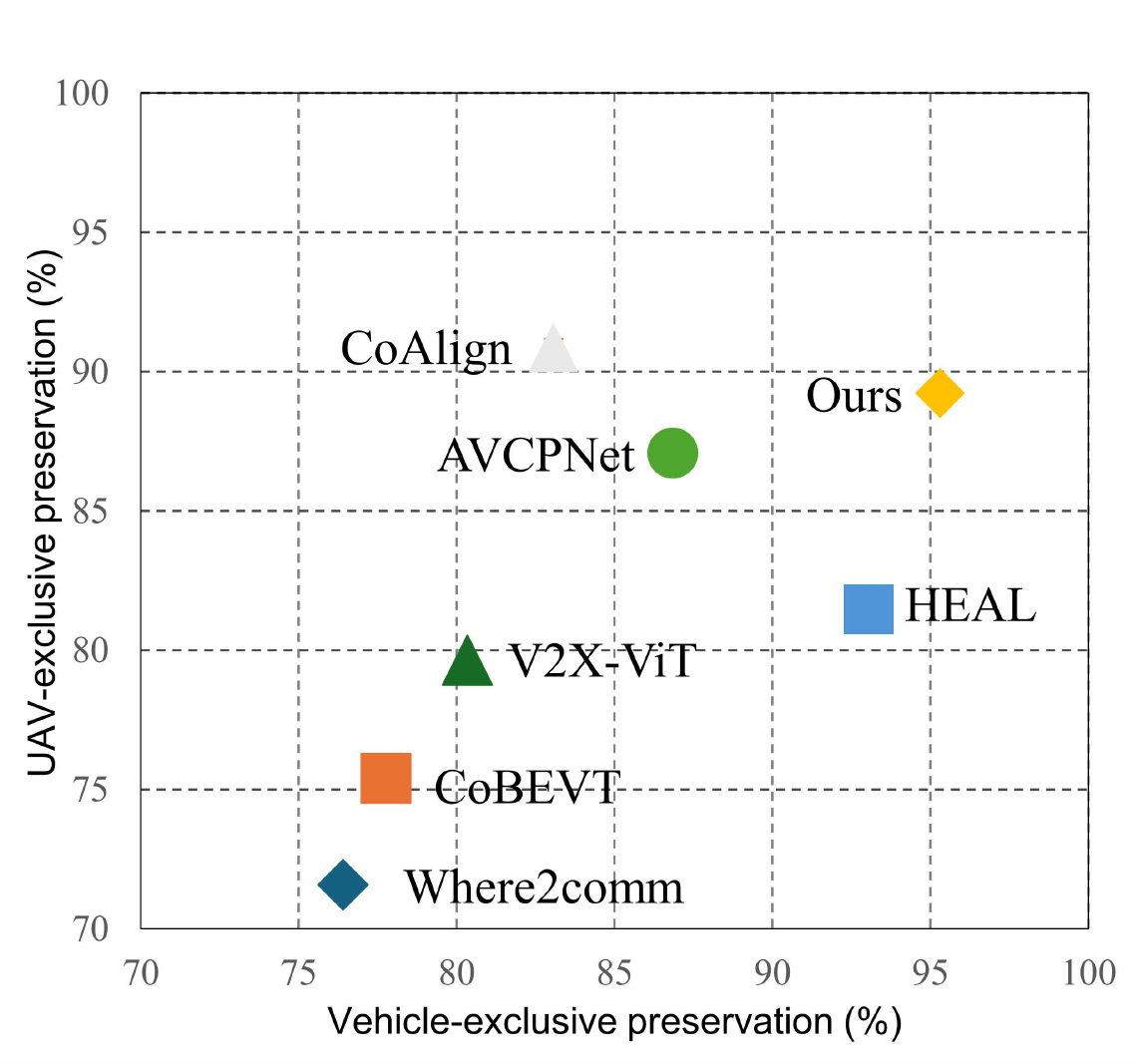}
    \caption{Preservation of agent-exclusive detection capabilities on OPV2V-Air. Each point represents a collaborative perception method. For each agent, we identify objects detected by that agent alone but missed by the other, using BEV IoU $\geq 0.7$ as the detection criterion. The preservation rate is the percentage of these objects that the collaborative method also detects at the same IoU threshold.}
    \label{fig:recall}
\end{figure}

\begin{figure}[t]
    \centering
    \begin{subfigure}[t]{0.49\columnwidth}
        \centering
        \includegraphics[width=\linewidth]{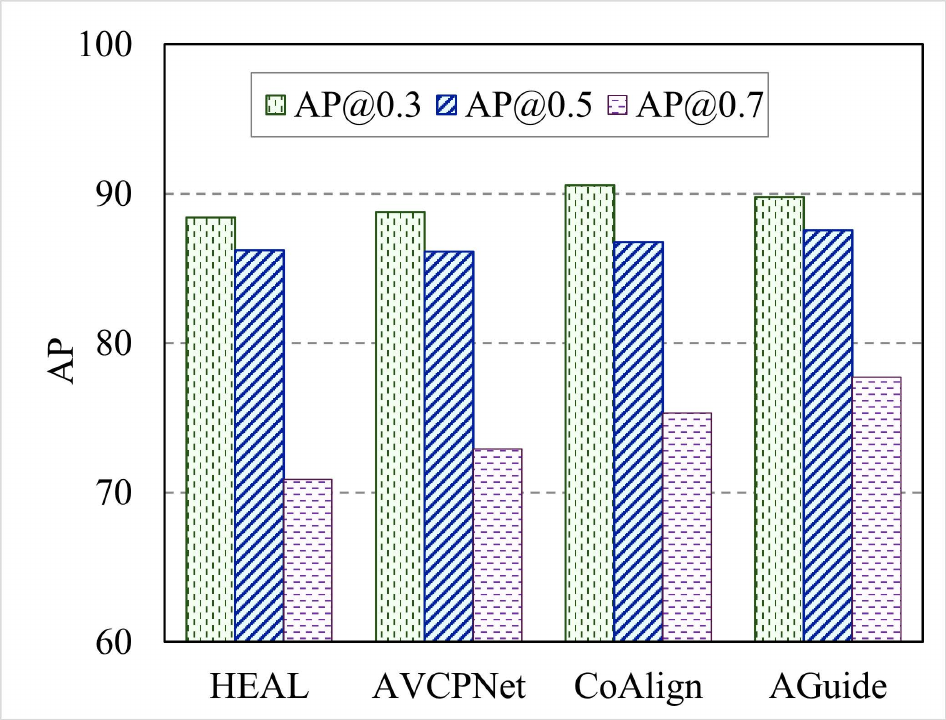}
        \caption{Near-to-mid range ($<50$\,m).}
        \label{fig:distance_lt50}
    \end{subfigure}
    \hfill
    \begin{subfigure}[t]{0.49\columnwidth}
        \centering
        \includegraphics[width=\linewidth]{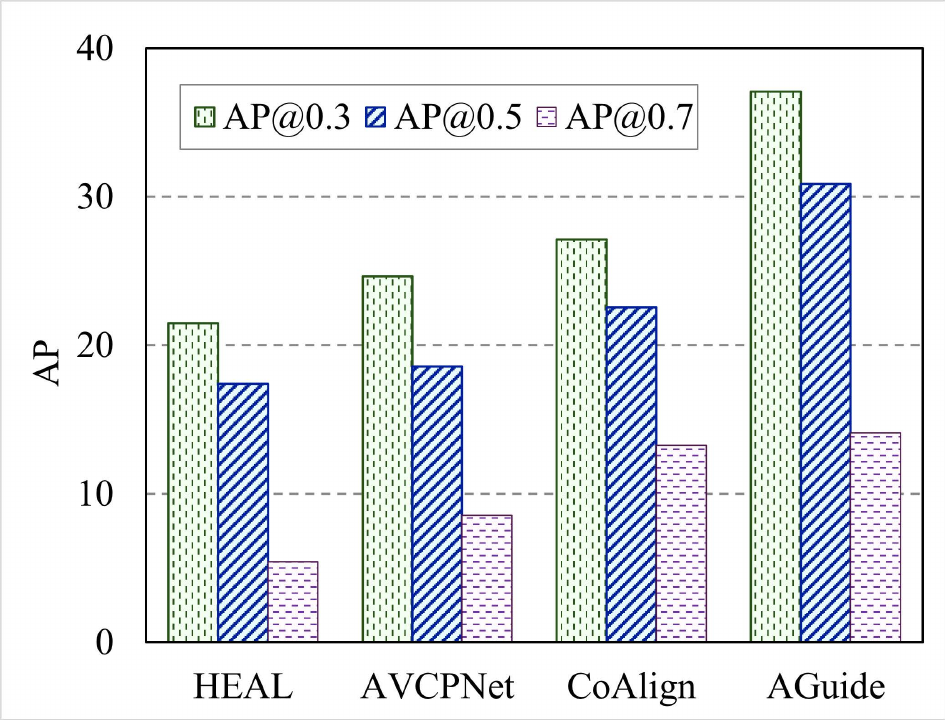}
        \caption{Long range ($\geq50$\,m).}
        \label{fig:distance_ge50}
    \end{subfigure}
    \caption{Distance-dependent detection performance on the
OPV2V-Air test set. Objects are grouped by their BEV center
distance to the ego vehicle, using 50\,m as the boundary.}
    \label{fig:distance}
\end{figure}

\begin{table}[t]
\centering
\small
\setlength{\tabcolsep}{4pt}
\renewcommand{\arraystretch}{1.08}
\caption{Detection performance under different collaboration
configurations on OPV2V-Air, reported as AP@0.3 / AP@0.5 / AP@0.7
(\%). With an additional vehicle, AGuide achieves the highest
AP across all three IoU thresholds, demonstrating its scalability
to a larger collaboration setting.}
\label{tab:collaboration_scalability}
\begin{tabular}{@{}lcc@{}}
\toprule
Method & 1 Vehicle + 1 UAV & 2 Vehicles + 1 UAV \\
\midrule
Vehicle(s) & 57.0 / 43.1 / 23.3 & 76.9 / 67.2 / 46.3 \\
UAV & 87.5 / 82.5 / 66.3 & 87.5 / 82.5 / 66.3 \\
Late Fusion & 78.0 / 74.4 / 61.2 & 89.5 / 84.0 / 66.2 \\
\midrule
CoBEVT & 81.2 / 77.0 / 59.4 & 83.8 / 80.7 / 65.2 \\
Where2comm & 85.1 / 79.0 / 58.8 & 87.1 / 82.5 / 64.0 \\
V2X-ViT & 82.7 / 78.4 / 62.2 & 85.5 / 82.1 / 68.0 \\
CoAlign & \textbf{87.8} / \underline{83.9} / \underline{72.4}
        & 89.8 / 86.2 / 75.5 \\
HEAL & 85.5 / 83.2 / 67.8
     & 90.4 / 88.5 / \underline{77.3} \\
AVCPNet$^\ddagger$ & 86.0 / 83.2 / 70.0
                  & \underline{91.3} / \underline{88.6} / 76.5 \\
AGuide (Ours) & \underline{87.6} / \textbf{85.1} / \textbf{74.9}
              & \textbf{92.3 / 90.3 / 82.5} \\
\bottomrule
\end{tabular}
\end{table}

\begin{figure*}[th]
    \centering
    \begin{subfigure}[t]{0.49\textwidth}
        \centering
        \includegraphics[
    width=\linewidth,
    height=4.2cm,
    keepaspectratio
]{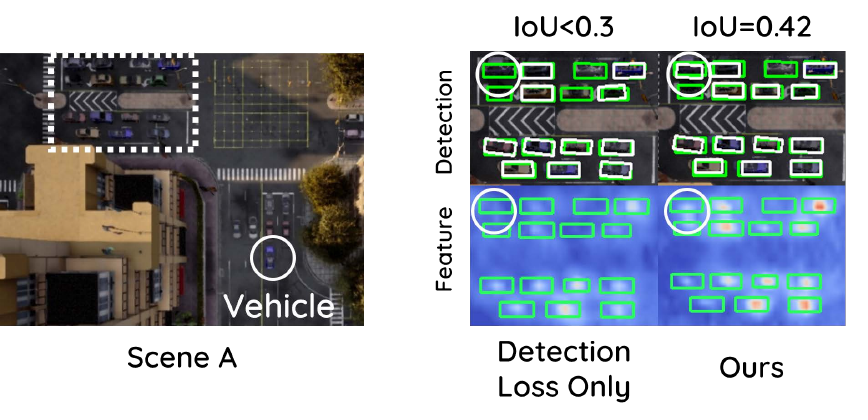}
        \caption{Long-range recovery.}
        \label{fig:long_range}
    \end{subfigure}
    \hfill
    \begin{subfigure}[t]{0.49\textwidth}
        \centering
        \includegraphics[
    width=\linewidth,
    height=4.2cm,
    keepaspectratio
]{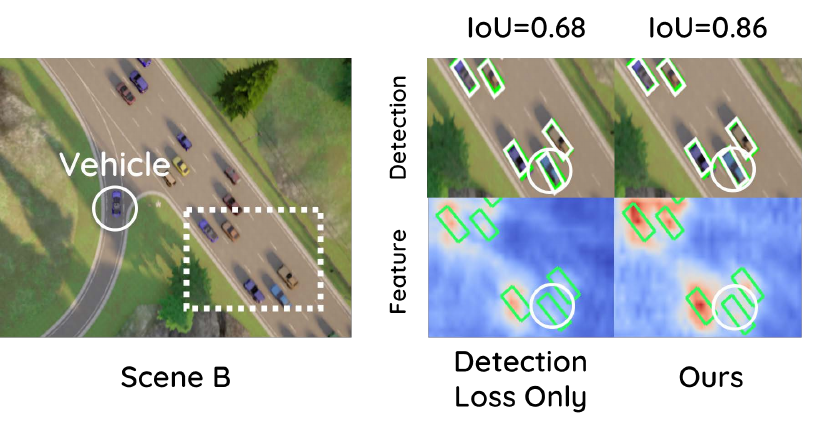}
        \caption{Mid-range refinement.}
        \label{fig:mid_range}
    \end{subfigure}
    \caption{Qualitative comparison between detection-only supervision
and AGuide on OPV2V-Air, illustrating long-range target recovery
in (a) and mid-range localization refinement in (b).
Dashed rectangles mark enlarged regions, while white circles
highlight selected targets. Green boxes denote ground truth,
including target locations on the feature maps, and white boxes
denote predictions.}
    \label{fig:recovery_refinement}
\end{figure*}
\begin{figure*}[t]
    \centering
    \includegraphics[width=\textwidth]{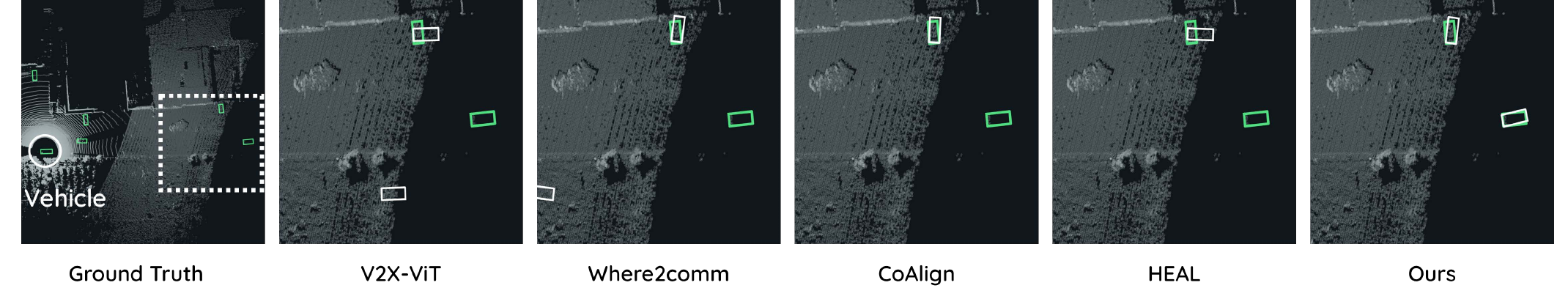}
    \caption{Qualitative comparison on V2U4Real with one vehicle and one UAV. AGuide recovers a distant target missed by the compared methods while producing fewer false positives in this scene. Green and white boxes denote ground truth and predictions, respectively.}
    \label{fig:v2u4real_vis}
\end{figure*}

\subsubsection{Analysis of Collaborative Gains}
\paragraph{Agent Capability Preservation.}
Figure~\ref{fig:recall} reveals different trade-offs in preserving agent-exclusive detections. Among vehicle--vehicle collaborative methods, CoBEVT, Where2comm, and V2X-ViT show relatively low preservation on both subsets, whereas CoAlign retains considerably more UAV-exclusive detections but remains weaker on vehicle-exclusive detections. The air--ground collaborative method AVCPNet also achieves high UAV-exclusive preservation but relatively low vehicle-exclusive preservation. In contrast, HEAL's vehicle-centered adaptation design preserves vehicle-exclusive detections well but retains fewer UAV-exclusive detections. Building on this design, AGuide introduces paired velocity prediction supervision for UAV adaptation, improving UAV-exclusive preservation while further enhancing vehicle-exclusive preservation.

\paragraph{Distance-Dependent Performance.}
As shown in Figure~\ref{fig:distance}, AGuide achieves the highest
AP@0.7 within 50\,m, indicating improved detection quality under
a strict localization criterion. Beyond 50\,m, its gains are more
pronounced at AP@0.3 and AP@0.5, suggesting better recovery of
distant objects. These results highlight its complementary
strengths in near-to-mid-range precise localization and long-range
object recovery, further illustrated by the qualitative examples.

\subsubsection{Scalability to Larger Collaboration Groups}

Table~\ref{tab:collaboration_scalability} evaluates the extension
from one-vehicle--one-UAV to two-vehicle--one-UAV collaboration.
AGuide leads in AP@0.5 and AP@0.7 in the smaller group and
across all three IoU thresholds in the larger group, whereas
CoAlign loses its second-place position in AP@0.5 and AP@0.7.
Maintaining this advantage as the group expands supports the
scalability of AGuide to more diverse collaboration configurations.

\subsection{Qualitative Results}

\paragraph{Long-Range Recovery and Mid-Range Refinement on OPV2V-Air.}
Figure~\ref{fig:recovery_refinement} shows how improved feature
responses relate to the detection gains of AGuide. For the distant
target, detection-only supervision yields a weak response, whereas
AGuide produces a more distinct activation around the target,
accompanied by successful recovery. For the mid-range target,
AGuide exhibits a more focused response and a predicted box that
better matches the ground truth. These examples complement the
quantitative results, illustrating the benefits of feature
adaptation for long-range object recovery and mid-range precise
localization.

\paragraph{Real-World Collaborative Detection on V2U4Real.}
Figure~\ref{fig:v2u4real_vis} presents a qualitative comparison
in a real-world LiDAR scene. AGuide detects the distant target
on the right that all compared methods miss, while avoiding
the spurious detections visible in several baseline predictions.
This example illustrates how its detection benefits extend
to real-world air--ground collaboration.
\section{Conclusion}

We presented AGuide, a framework that uses Flow Matching as auxiliary supervision to learn UAV representations for vehicle-side fusion and detection. Paired velocity prediction at object and token levels provides cross-platform guidance, using vehicle-teacher ROI references and deformable local references to capture regional semantics and accommodate local response differences. We also introduced OPV2V-Air, a camera-based simulation benchmark for air--ground collaborative perception. Experiments on OPV2V-Air and V2U4Real support the effectiveness of AGuide across camera and LiDAR inputs, with particular benefits for accurate localization. Future work will further investigate how Flow Matching guides feature adaptation and explore its application to heterogeneous collaborative perception and other tasks requiring cross-domain adaptation.

\section*{LLM Usage Statement}
We used ChatGPT (OpenAI) as a writing assistant during manuscript
preparation. It helped refine the grammar, wording, and clarity of
selected passages and draft this statement. We reviewed the revised
text for technical accuracy and take full responsibility for the
submitted manuscript.

{
    \small
    \bibliographystyle{ieeenat_fullname}
    \bibliography{main}
}

\clearpage
\setcounter{page}{1}
\maketitlesupplementary

\begin{figure*}[t]
    \centering
    \includegraphics[width=1\linewidth]{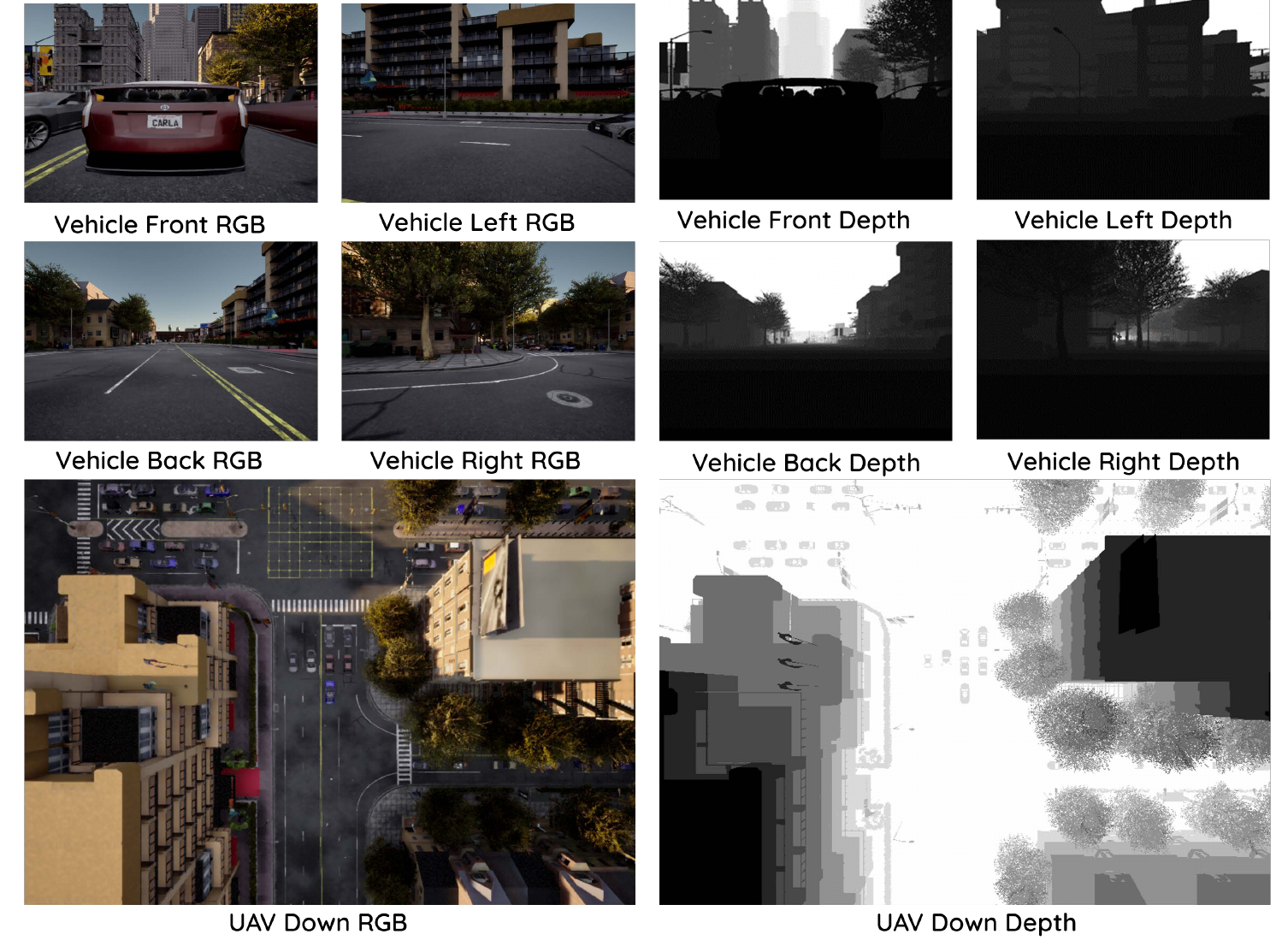}
    \caption{Synchronized RGB and depth observations from OPV2V-Air.
The top two rows show the front, left, rear, and right vehicle
views. The bottom row shows the downward-facing UAV view.}
\label{fig:opv2v_air_observations}
\end{figure*}

\section{Additional Details of OPV2V-Air}
\label{app:dataset_details}

\subsection{Scenario Construction}

We used a modified OpenCDA pipeline~\cite{xu2021opencda} to
replay 68 OPV2V scenarios~\cite{xu2022opv2v} in CARLA 0.9.12
across eight towns (Town01--Town07 and Town10).
At each replay step, we restored the recorded vehicle poses
and rendered the vehicle and UAV observations in the same
CARLA world coordinate system.
The simulator ran synchronously with a fixed time step of
$0.05$\,s, and observations were saved every other simulation
tick, yielding a nominal sampling rate of 10\,Hz.

\subsection{Sensor Configurations}

Each ground vehicle is equipped with four RGB cameras and four
corresponding depth cameras facing forward, left, right, and
rear. The nominal yaw angles are $0^\circ$, $100^\circ$,
$-100^\circ$, and $180^\circ$ relative to the vehicle,
respectively. The aerial sensor rig is positioned at a nominal
height of 50\,m above its associated ground vehicle and carries
one downward-facing RGB camera and one downward-facing depth
camera. In addition, each vehicle carries a 64-channel LiDAR and
each UAV carries a 32-channel LiDAR; Table~\ref{tab:sensor_config}
summarizes all sensor settings.

\begin{table}[h]
    \centering
    \small
    \caption{Sensor configurations of OPV2V-Air. Camera fields
    of view are horizontal, and LiDAR fields of view are vertical.}
    \label{tab:sensor_config}
    \setlength{\tabcolsep}{0pt}
    \renewcommand{\arraystretch}{1.05}
    \begin{tabular*}{\columnwidth}{@{\extracolsep{\fill}}lll@{}}
        \toprule
        Agent & Sensor & Configuration \\
        \midrule
        Vehicle & 4 RGB cameras   & $800\times600$, $100^\circ$ \\
        Vehicle & 4 depth cameras & $800\times600$, $100^\circ$ \\
        Vehicle & 64-channel LiDAR & $[-30^\circ,10^\circ]$ \\
        \midrule
        UAV & 1 RGB camera   & $800\times600$, $100^\circ$ \\
        UAV & 1 depth camera & $800\times600$, $100^\circ$ \\
        UAV & 32-channel LiDAR & $[-90^\circ,-30^\circ]$ \\
        \bottomrule
    \end{tabular*}
\end{table}

\subsection{Dataset Statistics and Splits}

OPV2V-Air comprises 68 scenarios, 35,313 UAV-view records, and 890,710 aerial-view 3D vehicle box annotations. We count each UAV view as a separate record and sum the box annotations across those records. The dataset contains 565,008 files and occupies approximately 221\,GB.

We split OPV2V-Air at the scenario level into 43 training, 9 validation, and 16 test scenarios. Table~\ref{tab:dataset_split} summarizes the UAV-view records and box annotations in each split. Although the splits differ in the number of UAV-view records and the density of vehicle annotations, we retain the original OPV2V scenario assignments with no additional rebalancing.

\begin{table}[t]
    \centering
    \small
    \caption{Scenario-level splits of OPV2V-Air. Box counts sum the 3D vehicle annotations across UAV-view records in each split.}
    \label{tab:dataset_split}
    \setlength{\tabcolsep}{0pt}
    \renewcommand{\arraystretch}{1.05}
    \begin{tabular*}{\columnwidth}{@{\extracolsep{\fill}}lrrr@{}}
        \toprule
        Split & Scenarios & UAV-view records & UAV boxes \\
        \midrule
        Training   & 43 & 19,123 & 332,627 \\
        Validation &  9 & 10,382 & 461,952 \\
        Test       & 16 &  5,808 &  96,131 \\
        \midrule
        Total      & 68 & 35,313 & 890,710 \\
        \bottomrule
    \end{tabular*}
\end{table}

\subsection{Annotations and Training Signals}

\paragraph{3D Vehicle Annotations.}
We derive 3D bounding boxes from CARLA actor states. For each
sensor rig, we consider other vehicles within 120\,m of its
associated ground vehicle and use instance returns from the
corresponding semantic LiDAR scan to select which vehicles
to annotate. Each annotation records the vehicle instance ID,
its world position and orientation, and the box center and
half-extents in the vehicle's local coordinate frame.

\paragraph{Training Signals.}
In our OPV2V-Air experiments, RGB images are the model inputs.
We use 3D boxes to supervise vehicle detection and UAV depth
maps to supervise aerial depth estimation. The dataset also
provides LiDAR point clouds and bird's-eye-view (BEV)
visibility maps.

\subsection{Example Observations}

Figure~\ref{fig:opv2v_air_observations} shows synchronized
observations from a ground vehicle and its associated UAV
at a single timestamp. The ground cameras capture the scene
from four directions at street level, while the downward-facing
UAV camera shows how the surrounding streets and vehicles are
arranged around the intersection.


\section{V2U4Real Evaluation Protocol}
\label{app:v2u4real_protocol}

The complete V2U4Real test split is not publicly available, so we use the public validation split for evaluation. We use pretrained checkpoints from the official V2U4Real repository where available. For baselines without released checkpoints, we train them on the public training split using the same LiDAR inputs and one-vehicle, one-UAV setting as AGuide.

\section{Additional Implementation Details}
\label{app:implementation_details}

\subsection{Training Protocol}

AGuide follows a unified three-stage training protocol across both benchmarks, adapted only to each dataset's native sensor modality and available agent configuration.

\paragraph{Stage~I: Vehicle Teacher Training.}
We first train the vehicle teacher to establish the reference representation space, fusion module, and detection head, and then freeze it for subsequent stages. On OPV2V-Air, we train with one ego vehicle and one cooperating vehicle ($x,y\in[-51.2,51.2]$\,m, $z\in[-3,1]$\,m) so that Pyramid Fusion learns foreground estimation for multi-vehicle fusion. On V2U4Real, where each scene contains a single ground vehicle, the vehicle teacher is trained in the single-vehicle setting.

\paragraph{Stage~II: UAV Representation Adaptation.}
We initialize the UAV branch from the Stage~I model and adapt it using observations from one UAV and references extracted by the frozen vehicle teacher. We update the UAV encoder, BEV backbone, adapter, pose modulation module, and auxiliary velocity networks while keeping Pyramid Fusion and the detection head frozen. Both benchmarks use detection and Flow Matching supervision, with auxiliary UAV depth supervision additionally included for camera-based BEV lifting on OPV2V-Air.

\paragraph{Stage~III: Fusion Tuning.}
We integrate the adapted UAV branch with the Stage~I vehicle branch and fine-tune the UAV adapter and the final layer of the pose modulation module using the collaborative detection loss. Fine-tuning uses two vehicles and one UAV on OPV2V-Air (retaining its Stage~I multi-vehicle configuration) and one vehicle and one UAV on V2U4Real.

\paragraph{Modality and Inference Configurations.}
Because V2U4Real provides LiDAR point clouds rather than RGB images, each agent extracts BEV features directly from point clouds, naturally omitting the camera-based image-to-BEV projection (including GBG) and depth supervision. For inference, our main experiments evaluate one ego vehicle and one UAV on both benchmarks for a consistent cross-dataset protocol. We additionally report two-vehicle, one-UAV results on OPV2V-Air in Table~\ref{tab:collaboration_scalability} of the main paper.

\subsection{Auxiliary Supervision Settings}

\paragraph{Object-Level Flow Matching.}
We retain vehicle teacher proposals with a confidence score
of at least $0.3$ and a BEV IoU of at least $0.5$ with a
ground-truth box. We use ROIAlign to extract $3\times3$ features
and apply global average pooling to obtain object prototypes.

\paragraph{Token-Level Flow Matching.}
We partition the BEV feature maps into $3\times3$ patches.
The deformable sampler uses nine sampling points per token
and a configured radius of two cells on the token grid.
The token loss assigns weights of $1.5$ to regions covered
by reliable teacher proposals, $0.2$ to regions covered by
ground-truth boxes but no reliable teacher proposal, and
$0.02$ to background. For background tokens, we use the
vehicle token at the same grid location as the reference.

\paragraph{Loss Weights.}
In Stage II on OPV2V-Air, the UAV depth loss has a weight of
$0.2$, and the auxiliary feature adaptation objective has
an overall weight of $0.03$. Within the auxiliary objective,
the token-level and object-level Flow Matching losses have
relative weights of $1.0$ and $1.2$, respectively.

\section{Representation Analysis Metrics}
\label{app:representation_metrics}

For representation analysis on the OPV2V-Air test set, all methods are evaluated on the same ground-truth boxes. For each box, we apply ROIAlign and global average pooling to the adapted UAV and frozen vehicle BEV feature maps to obtain paired object features $\mathbf{u}_i$ and $\mathbf{v}_i$.

\subsection{Squared Maximum Mean Discrepancy}
\label{app:mmd_metric}

We use squared maximum mean discrepancy ($\mathrm{MMD}^{2}$) to compare the distributions of adapted UAV and frozen vehicle object features. Each feature vector is $\ell_2$-normalized before computation. Let $N$ be the number of sampled objects. We compute the biased empirical estimate of squared MMD as
\begin{equation}
\begin{aligned}
\mathrm{MMD}^{2}
={}& \frac{1}{N^{2}}\sum_{i,j=1}^{N}
\bigl[k(\mathbf{u}_i,\mathbf{u}_j)+k(\mathbf{v}_i,\mathbf{v}_j)\bigr] \\
&-\frac{2}{N^{2}}\sum_{i,j=1}^{N} k(\mathbf{u}_i,\mathbf{v}_j),
\end{aligned}
\end{equation}
where $k$ is a radial basis function kernel,
\begin{equation}
k(\mathbf{x},\mathbf{y})
=\exp\left(-\frac{\|\mathbf{x}-\mathbf{y}\|_2^2}{2\sigma^2}\right).
\end{equation}
We set the bandwidth $\sigma$ to the median pairwise Euclidean distance in a common pool of reference features and use the same bandwidth for all methods. Lower $\mathrm{MMD}^{2}$ indicates closer feature distributions under this kernel. The estimator compares all feature pairs across the two distributions, independently of the object correspondences used for feature extraction.

\subsection{Directional Linear Predictability}
\label{app:directional_r2}

While $\mathrm{MMD}^{2}$ is symmetric and unpaired, we further evaluate directional linear predictability ($R^{2}$) between paired UAV and frozen vehicle ROI features $(\mathbf{u}_i, \mathbf{v}_i)$. Following the standalone expert matching criterion at BEV IoU $\ge 0.7$ (Table~\ref{tab:exclusive_preservation}), test objects are partitioned into Both (detected by both experts) and UAV-only (detected only by the UAV). Within each subset, we fit ridge regression to predict $\mathbf{v}_i$ from $\mathbf{u}_i$ ($\mathrm{U} \rightarrow \mathrm{V}$) and vice versa ($\mathrm{V} \rightarrow \mathrm{U}$), computing
\begin{equation}
\begin{aligned}
R^{2}_{\mathrm{U}\rightarrow\mathrm{V}}
&= 1 - \frac{\sum_{i}\|\mathbf{v}_i - \hat{\mathbf{v}}_i\|_2^2}{\sum_{i}\|\mathbf{v}_i - \bar{\mathbf{v}}\|_2^2}, \\
R^{2}_{\mathrm{V}\rightarrow\mathrm{U}}
&= 1 - \frac{\sum_{i}\|\mathbf{u}_i - \hat{\mathbf{u}}_i\|_2^2}{\sum_{i}\|\mathbf{u}_i - \bar{\mathbf{u}}\|_2^2},
\end{aligned}
\end{equation}
where $\hat{\mathbf{v}}_i$ and $\hat{\mathbf{u}}_i$ denote the ridge-regression predictions, and $\bar{\mathbf{v}}$ and $\bar{\mathbf{u}}$ denote the corresponding subset feature means. Since $\mathbf{v}_i$ is frozen across all variants, shifts relative to Detection Only isolate how auxiliary supervision reshapes UAV features: higher $R^{2}_{\mathrm{U}\rightarrow\mathrm{V}}$ reflects stronger decodability into the vehicle space, whereas elevated $R^{2}_{\mathrm{V}\rightarrow\mathrm{U}}$---especially on UAV-only objects---indicates stronger convergence toward incomplete vehicle representations.

\section{Additional Experimental Results}

\subsection{Sensitivity to Random Seeds}

Table~\ref{tab:seed_sensitivity} reports the performance of AGuide
with three random seeds on the OPV2V-Air test split with one ego
vehicle and one UAV. AP@0.7 ranges from 74.46 to 74.90 across the
three runs, with a sample standard deviation of 0.24 percentage
points.

\begin{table}[h]
    \centering
    \small
    \caption{Detection performance of AGuide across three random
    seeds on OPV2V-Air. AP is reported in percent. The last row
    reports the mean and sample standard deviation.}
    \label{tab:seed_sensitivity}
    \begin{tabular}{@{}lccc@{}}
        \toprule
        Seed & AP@0.3 & AP@0.5 & AP@0.7 \\
        \midrule
        303 & 87.60 & 85.10 & 74.90 \\
        111 & 88.23 & 85.56 & 74.46 \\
        222 & 87.40 & 85.03 & 74.50 \\
        \midrule
        Mean $\pm$ SD & $87.74 \pm 0.43$ & $85.23 \pm 0.29$ & $74.62 \pm 0.24$ \\
        \bottomrule
    \end{tabular}
\end{table}

\subsection{Supplementary Ablation on Feature Supervision}

We compared Full FM with direct feature alignment using cosine distance. Endpoint Cosine retained the reliable ROI and deformable token correspondences and replaced velocity regression in both branches with weighted cosine distance between the UAV features and their vehicle references. As shown in Table~\ref{tab:supervision_cosine}, Endpoint Cosine increased AP@0.7 from 71.8 to 73.7 compared with Detection Only, while AP@0.3 and AP@0.5 decreased slightly. Full FM achieved the highest AP at all three IoU thresholds, reaching 74.9 at IoU 0.7. These results suggest that direct feature alignment can improve strict-IoU detection and that FM provides a further gain under the evaluated settings.

\begin{table}[t]
\caption{Comparison of supervision strategies on the OPV2V-Air test split. }
\centering
\small
\setlength{\tabcolsep}{4pt}
\begin{tabular}{@{}lccc@{}}
\toprule
Supervision & AP@0.3 & AP@0.5 & AP@0.7 \\
\midrule
Detection Only  & 86.9 & 84.3 & 71.8 \\
Endpoint Cosine & 86.5 & 83.9 & 73.7 \\
Full FM         & \textbf{87.6} & \textbf{85.1} & \textbf{74.9} \\
\bottomrule
\end{tabular}

\label{tab:supervision_cosine}
\end{table}


\subsection{Directional Feature Predictability Analysis}
\label{sec:directional_r2}

Using the directional linear predictability metric defined in Section~\ref{app:directional_r2}, Table~\ref{tab:directional_r2} compares how different auxiliary supervision strategies shape cross-platform ROI representations on the OPV2V-Air test split. Endpoint MSE increases predictability in both directions across the two subsets and produces the largest $\mathrm{V}\!\to\!\mathrm{U}$ shift on UAV-only objects, consistent with its degraded AP@0.7. Endpoint Cosine avoids this severe drop, yet yields more limited $\mathrm{U}\!\to\!\mathrm{V}$ gains on UAV-only objects while still noticeably raising $\mathrm{V}\!\to\!\mathrm{U}$. In contrast, Full FM markedly improves $\mathrm{U}\!\to\!\mathrm{V}$ predictability across both subsets while keeping $\mathrm{V}\!\to\!\mathrm{U}$ closest to the Detection Only baseline, attaining the highest AP@0.7. Together, these trends suggest that paired velocity supervision enhances vehicle-space decodability while better preserving UAV-specific information under asymmetric air-ground visibility.

\begin{table*}[t]
  \centering
  \caption{Directional linear predictability ($R^2$) across auxiliary supervision strategies on the OPV2V-Air test split (see Section~\ref{app:directional_r2}). Both and UAV-only denote co-visible objects and objects detected only by the UAV, respectively; $\mathrm{U}\!\to\!\mathrm{V}$ and $\mathrm{V}\!\to\!\mathrm{U}$ indicate linear prediction between adapted UAV and frozen vehicle ROI features. Numbers in parentheses denote changes relative to Detection Only. Only the highest AP@0.7 is in bold, as higher $R^2$ does not necessarily indicate better adaptation.}
  \label{tab:directional_r2}
  \small
  \setlength{\tabcolsep}{7pt}
  \begin{tabular}{lccccc}
    \toprule
    \multirow{2}{*}{Supervision} & \multirow{2}{*}{AP@0.7 (\%)} & \multicolumn{2}{c}{Both ($R^2$)} & \multicolumn{2}{c}{UAV-only ($R^2$)} \\
    \cmidrule(lr){3-4} \cmidrule(lr){5-6}
     & & $\mathrm{U}\!\to\!\mathrm{V}$ & $\mathrm{V}\!\to\!\mathrm{U}$ & $\mathrm{U}\!\to\!\mathrm{V}$ & $\mathrm{V}\!\to\!\mathrm{U}$ \\
    \midrule
    Detection Only  & 71.8 & 0.319 & 0.469 & 0.152 & 0.307 \\
    \midrule
    Endpoint MSE    & 68.9 & 0.443 {\scriptsize\textcolor{gray}{(+0.124)}} & 0.526 {\scriptsize\textcolor{gray}{(+0.057)}} & 0.205 {\scriptsize\textcolor{gray}{(+0.053)}} & 0.359 {\scriptsize\textcolor{gray}{(+0.052)}} \\
    Endpoint Cosine & 73.7 & 0.372 {\scriptsize\textcolor{gray}{(+0.053)}} & 0.488 {\scriptsize\textcolor{gray}{(+0.019)}} & 0.174 {\scriptsize\textcolor{gray}{(+0.022)}} & 0.337 {\scriptsize\textcolor{gray}{(+0.030)}} \\
    Full FM         & \textbf{74.9} & 0.380 {\scriptsize\textcolor{gray}{(+0.061)}} & 0.477 {\scriptsize\textcolor{gray}{(+0.008)}} & 0.196 {\scriptsize\textcolor{gray}{(+0.044)}} & 0.317 {\scriptsize\textcolor{gray}{(+0.010)}} \\
    \bottomrule
  \end{tabular}
\end{table*}

\subsection{Sensitivity to Vehicle-Side ROI Reference Shifts}

We isolate the effect of vehicle-side ROI shifts using the ROI-only variant of AGuide, which applies Flow Matching exclusively to object-level ROI prototypes. During Stage~II training, we select reliable proposals and then shift the vehicle-side ROI centers in random BEV directions before ROIAlign, keeping their paired UAV ROIs fixed. Table~\ref{tab:roi_perturbation} reports detection performance on the OPV2V-Air test split with one vehicle and one UAV, evaluated before Stage~III fine-tuning. A 0.25\,m shift has little effect on AP@0.3 and AP@0.5 but noticeably reduces AP@0.7. At 0.50\,m, all three AP metrics decline, with the largest drop at IoU~0.7.

\begin{table}[h]
\centering
\small
\caption{Effect of vehicle-side ROI center shifts during Stage~II training with ROI-only Flow Matching. Results are reported on the OPV2V-Air test split with one vehicle and one UAV before Stage~III fine-tuning. AP is reported in percent.}
\label{tab:roi_perturbation}
\begin{tabular}{@{}lccc@{}}
\toprule
ROI Center Offset & AP@0.3 & AP@0.5 & AP@0.7 \\
\midrule
0.00\,m & \textbf{86.14} & \textbf{83.60} & \textbf{69.36} \\
0.25\,m & 86.09 & 83.46 & 68.31 \\
0.50\,m & 84.24 & 81.63 & 66.91 \\
\bottomrule
\end{tabular}
\end{table}


\subsection{Flow Matching Ablation on V2U4Real}
We compare configurations with and without Flow Matching supervision
on the V2U4Real validation split using one vehicle and one UAV.
As shown in Table~\ref{tab:v2u4real_supervision}, the configuration
with Flow Matching achieved higher AP at all three IoU thresholds.

\begin{table}[h]
\centering
\small
\setlength{\tabcolsep}{4pt}
\renewcommand{\arraystretch}{1.08}
\caption{Ablation of Flow Matching supervision on V2U4Real.
AP is reported in \%.}
\label{tab:v2u4real_supervision}
\begin{tabular}{@{}lccc@{}}
\toprule
Configuration & AP@0.3 & AP@0.5 & AP@0.7 \\
\midrule
w/o Flow Matching & 56.5 & 55.9 & 36.6 \\
With Flow Matching & \textbf{62.9} & \textbf{61.7} & \textbf{40.6} \\
\bottomrule
\end{tabular}
\end{table}

\subsection{Effect of Cross-Platform Reference Pairing}

We examine how the choice of vehicle references affects detection before Stage~III fine-tuning. Within each minibatch, we replace the vehicle reference token map for each UAV frame with that of another frame, preserving the spatial indices. For the object branch, we draw each vehicle ROI reference from another frame in the same minibatch. If the batch has valid proposals in only one frame, we retain the original ROI pairs. Each substituted reference defines both the interpolation state and the target velocity.

Table~\ref{tab:stage2_reference_pairing} shows that correctly paired references achieve the highest AP at all three IoU thresholds. The largest gap between the two Flow Matching variants appears at IoU~0.7, where correct pairing clearly outperforms detection-only training while shuffling falls below it. This pattern is consistent with matched vehicle references being particularly useful for precise localization.

\begin{table}[h]
\centering
\small
\setlength{\tabcolsep}{4pt}
\renewcommand{\arraystretch}{1.08}
\caption{Stage~II comparison of Flow Matching reference pairing on OPV2V-Air. We combine each Stage~II checkpoint with the same Stage~I vehicle weights and evaluate with one vehicle and one UAV before Stage~III fine-tuning. All AP values are percentages.}
\label{tab:stage2_reference_pairing}
\begin{tabular}{@{}lccc@{}}
\toprule
Supervision & AP@0.3 & AP@0.5 & AP@0.7 \\
\midrule
Detection Only & 84.75 & 82.23 & 68.12 \\
FM (Shuffled References) & 83.91 & 81.57 & 67.38 \\
FM (Correctly Paired) & \textbf{85.51} & \textbf{83.12} & \textbf{70.33} \\
\bottomrule
\end{tabular}
\end{table}


\subsection{Velocity Direction Alignment}
Table~\ref{tab:velocity_alignment} reports the mean cosine similarity between predicted velocities and target displacements at fixed interpolation times on the OPV2V-Air test split. At $t=0$, the interpolated input contains only the UAV representation. Even at this endpoint, the mean cosine similarity reaches 0.790 for object prototypes and 0.924 for foreground tokens, showing that the predicted velocities are already aligned with the target directions on average. Alignment peaks at intermediate times across all three groups, where the input also contains information from the vehicle reference. This diagnostic characterizes how well the auxiliary networks recover the direction of the paired feature displacement at different points along the interpolation path.

\begin{table}[h]
\centering
\caption{Velocity direction alignment on the OPV2V-Air test split. We report the mean cosine similarity between predicted velocities and target feature displacements at fixed interpolation times. Object denotes the ROI-prototype branch; token results are reported separately for foreground and background.}
\small
\setlength{\tabcolsep}{6pt}
\renewcommand{\arraystretch}{1.08}
\begin{tabular}{@{}lccc@{}}
\toprule
$t$ & Object & Foreground Token & Background Token \\
\midrule
0.00 & 0.790 & 0.924 & 0.826 \\
0.25 & 0.926 & 0.980 & 0.949 \\
0.50 & 0.946 & 0.993 & 0.979 \\
0.75 & 0.929 & 0.994 & 0.984 \\
1.00 & 0.831 & 0.986 & 0.975 \\
\bottomrule
\end{tabular}
\label{tab:velocity_alignment}
\end{table}

\begin{table*}[t]
\centering
\small
\setlength{\tabcolsep}{5pt}
\renewcommand{\arraystretch}{1.10}
\caption{Preservation of agent-exclusive detections on OPV2V-Air (test) and V2U4Real (validation) with one vehicle and one UAV. Each entry reports recall at BEV IoU thresholds of 0.3, 0.5, and 0.7, respectively (\%). The Vehicle-only and UAV-only groups are fixed using matches from the standalone experts at IoU $\geq 0.7$. Each method is evaluated at its original operating point. Bold and underlining indicate the best and second-best values within each group and threshold at the displayed precision. $\ddagger$ denotes our reimplementation.}
\label{tab:exclusive_preservation}
\begin{tabular}{@{}lcccc@{}}
\toprule
& \multicolumn{2}{c}{OPV2V-Air} & \multicolumn{2}{c}{V2U4Real} \\
\cmidrule(lr){2-3}\cmidrule(l){4-5}
Method
& \shortstack{Vehicle-only \\ ($N=661$)}
& \shortstack{UAV-only \\ ($N=12{,}474$)}
& \shortstack{Vehicle-only \\ ($N=1{,}708$)}
& \shortstack{UAV-only \\ ($N=235$)} \\
\midrule
Where2comm & 91.5 / 89.3 / 76.4 & 97.2 / 92.3 / 71.6 & 96.7 / 96.7 / 85.6 & \underline{86.4 / 86.4} / 60.9 \\
CoBEVT & 94.6 / 92.0 / 77.8 & 94.2 / 92.1 / 75.4 & 95.3 / 94.6 / 78.2 & \underline{86.4} / 86.0 / \textbf{72.8} \\
V2X-ViT & 91.5 / 89.6 / 80.3 & 96.1 / 93.3 / 79.7 & 96.6 / 96.4 / 84.5 & 78.7 / 77.0 / 27.2 \\
CoAlign & 95.6 / 90.6 / 83.1 & \textbf{99.5} / \underline{98.3} / \textbf{90.9} & \textbf{98.2 / 98.2 / 91.2} & 84.3 / 84.3 / 43.8 \\
HEAL & 96.7 / 95.9 / \underline{93.0} & 98.5 / 97.6 / 81.5 & 93.4 / 93.3 / 87.1 & 72.3 / 72.3 / 46.0 \\
AVCPNet$^\ddagger$ & \underline{97.3 / 96.1} / 86.8 & 98.7 / 98.0 / 87.1 & 96.7 / 96.7 / 82.7 & 85.1 / 85.1 / 66.0 \\
AGuide (Ours) & \textbf{99.7 / 99.2 / 95.3} & \underline{99.0} / \textbf{98.4} / \underline{89.2} & \underline{97.8 / 97.7 / 91.0} & \textbf{88.9 / 88.9} / \underline{67.2} \\
\bottomrule
\end{tabular}
\end{table*}

\subsection{Preservation of Agent-Exclusive Detections}
Table~\ref{tab:exclusive_preservation} extends the main-text analysis to V2U4Real and evaluates recall at three IoU thresholds on the same expert-defined groups. On OPV2V-Air, AGuide achieves the highest recall on vehicle-only objects at all three thresholds and retains 89.2\% of UAV-only objects at IoU~0.7, second to CoAlign. On V2U4Real, AGuide ranks second on vehicle-only objects at all three thresholds and achieves the highest UAV-only recall at IoU~0.3 and 0.5. At IoU~0.7, CoBEVT leads on UAV-only objects, while AGuide ranks second. Across both datasets, AGuide retains a large share of detections unique to either platform, with the relative ranking depending on the IoU threshold and expert group.

\subsection{Detection Performance by Distance}
Tables~\ref{tab:distance_ap_10m_air} and~\ref{tab:distance_ap_10m_real} report BEV AP at IoU thresholds of 0.3, 0.5, and 0.7 on OPV2V-Air and V2U4Real, respectively. Results are grouped into 10\,m bins by distance from the ego vehicle. We include the number of ground-truth objects in each bin to show how samples are distributed across distances. AGuide achieves competitive detection performance across all reported distance bins on both OPV2V-Air and V2U4Real.

\begin{table*}[t]
\centering
\small
\setlength{\tabcolsep}{4pt}
\renewcommand{\arraystretch}{1.10}
\caption{Distance-dependent detection on the OPV2V-Air test split. Each entry reports BEV AP at IoU thresholds of 0.3, 0.5, and 0.7, respectively (\%). Distance bins are relative to the ego vehicle.  $\ddagger$ denotes our reimplementation.}
\label{tab:distance_ap_10m_air}
\resizebox{\textwidth}{!}{%
\begin{tabular}{@{}lccccccc@{}}
\toprule
Method & 0--10\,m & 10--20\,m & 20--30\,m & 30--40\,m & 40--50\,m & 50--60\,m & 60--70\,m \\
GT count & 2,066 & 5,248 & 5,165 & 5,525 & 4,798 & 994 & 137 \\
\midrule
No Fusion (Veh.) & 91.7 / 89.5 / 75.2 & 85.6 / 77.3 / 44.9 & 54.1 / 35.3 / 12.3 & 25.3 / 11.7 / 2.2 & 7.2 / 2.0 / 0.2 & 2.1 / 0.5 / 0.0 & \underline{5.7} / \underline{2.5} / 0.1 \\
No Fusion (UAV) & \underline{96.9} / 94.1 / 82.5 & 95.0 / 91.4 / 77.9 & 92.3 / 87.0 / 72.1 & 83.3 / 79.6 / 64.7 & 68.9 / 60.8 / 40.4 & 34.5 / \underline{29.8} / \textbf{17.0} & 1.8 / 0.6 / 0.1 \\
Late Fusion & 87.9 / 87.2 / 79.6 & 86.0 / 84.2 / 74.3 & 86.0 / 82.4 / 69.3 & 80.9 / 77.6 / 63.2 & \underline{69.0} / \underline{61.2} / 40.5 & \underline{35.0} / \underline{29.8} / \textbf{17.0} & 4.8 / 1.9 / \underline{0.2} \\
\midrule
CoBEVT & 91.9 / 91.5 / 82.8 & 94.2 / 93.1 / 81.0 & 88.1 / 87.0 / 70.5 & 78.5 / 74.4 / 53.2 & 60.6 / 50.5 / 27.9 & 21.8 / 17.5 / 8.6 & 2.0 / 0.5 / 0.0 \\
Where2comm & 94.8 / 92.6 / 80.9 & 95.4 / 92.9 / 82.4 & 91.7 / 88.5 / 69.0 & 81.9 / 75.6 / 47.1 & 63.6 / 52.0 / 28.0 & 16.8 / 10.3 / 4.3 & 0.1 / 0.0 / 0.0 \\
V2X-ViT & 96.4 / \underline{95.6} / \underline{90.7} & 95.1 / 93.7 / 83.5 & 91.1 / 88.9 / 73.2 & 77.0 / 70.8 / 51.3 & 61.8 / 54.1 / 34.6 & 24.0 / 20.4 / 13.0 & 0.5 / 0.2 / 0.0 \\
CoAlign & \textbf{97.5} / \textbf{97.1} / \textbf{93.3} & 96.0 / 94.0 / 86.1 & 94.1 / 92.7 / \underline{84.8} & 85.7 / 82.9 / \underline{69.2} & \textbf{71.1} / \textbf{62.3} / \textbf{44.2} & 30.1 / 25.4 / 15.0 & 2.8 / 1.2 / \textbf{0.5} \\
HEAL & 93.6 / 93.0 / 90.5 & 96.2 / 95.9 / 88.4 & 93.2 / 92.6 / 77.4 & 86.1 / 85.3 / 63.4 & 62.2 / 55.5 / 32.2 & 26.2 / 21.4 / 6.5 & 0.6 / 0.2 / 0.0 \\
AVCPNet$^\ddagger$ & 91.2 / 91.2 / 88.9 & \textbf{98.0} / \textbf{97.8} / \textbf{94.1} & \underline{94.3} / \underline{93.9} / 81.4 & \underline{88.0} / \underline{86.7} / 68.4 & 67.4 / 58.3 / 33.8 & 29.0 / 21.9 / 10.0 & 0.1 / 0.0 / 0.0 \\
AGuide (Ours) & 91.8 / 91.4 / 89.9 & \underline{96.6} / \underline{96.4} / \underline{92.6} & \textbf{95.0} / \textbf{94.3} / \textbf{85.9} & \textbf{89.7} / \textbf{88.8} / \textbf{75.5} & 67.6 / 60.3 / \underline{41.5} & \textbf{39.5} / \textbf{34.2} / \underline{16.2} & \textbf{11.1} / \textbf{3.8} / 0.1 \\
\bottomrule
\end{tabular}%
}
\end{table*}

\begin{table*}[t]
\centering
\small
\setlength{\tabcolsep}{4pt}
\renewcommand{\arraystretch}{1.10}
\caption{Distance-dependent detection on the V2U4Real validation split. Each entry reports BEV AP at IoU thresholds of 0.3, 0.5, and 0.7, respectively (\%). Distance bins are relative to the ego vehicle. $\ddagger$ denotes our reimplementation. }
\label{tab:distance_ap_10m_real}
\begin{tabular}{@{}lccccc@{}}
\toprule
Method & 0--10\,m & 10--20\,m & 20--30\,m & 30--40\,m & 40--50\,m \\
GT count & 2,799 & 372 & 821 & 1,179 & 507 \\
\midrule
No Fusion (Veh.) & 97.1 / \textbf{97.1} / 66.6 & 37.2 / 37.2 / 20.3 & 13.2 / 12.8 / 10.4 & 10.2 / 9.5 / 4.7 & 17.2 / 15.5 / 7.5 \\
No Fusion (UAV) & 59.2 / 58.0 / 34.4 & 11.7 / 6.4 / 1.6 & 4.0 / 2.4 / 0.0 & 5.1 / 3.6 / 0.5 & 3.7 / 1.2 / 0.4 \\
Late Fusion & 94.5 / 93.1 / 51.2 & 32.0 / 29.7 / 12.0 & 15.2 / 13.9 / 7.5 & 10.5 / 9.0 / 2.8 & 18.2 / 14.3 / 6.4 \\
\midrule
CoBEVT & 95.5 / 95.1 / 63.3 & 28.9 / 26.5 / 4.4 & 10.6 / 10.1 / 3.6 & 12.1 / 10.8 / 3.2 & 22.4 / 16.0 / 7.0 \\
Where2comm & 96.9 / 96.5 / 60.5 & 51.4 / 50.7 / 37.2 & 49.8 / 45.3 / 18.2 & \underline{32.6} / 27.8 / 10.5 & 26.2 / 21.7 / 12.4 \\
V2X-ViT & 96.3 / 95.3 / 57.3 & 36.8 / 34.1 / 21.0 & 8.2 / 7.3 / 5.3 & 9.9 / 9.3 / 4.3 & 17.4 / 15.7 / 8.3 \\
CoAlign & 97.1 / \textbf{97.1} / 64.2 & 61.7 / 61.7 / \underline{55.4} & \underline{49.9} / \underline{49.8} / \textbf{36.8} & 30.1 / \underline{28.5} / \textbf{17.0} & \underline{32.1} / 24.4 / \underline{13.4} \\
HEAL & 96.5 / 96.4 / \textbf{74.7} & 32.9 / 32.9 / 14.3 & 12.8 / 12.5 / 8.8 & 9.9 / 9.3 / 5.0 & 15.1 / 14.8 / 8.4 \\
AVCPNet$^\ddagger$ & \textbf{97.3} / \underline{96.7} / 71.5 & \underline{75.8} / \underline{75.8} / 55.0 & 23.1 / 23.1 / 17.0 & 14.7 / 14.7 / 8.7 & 30.7 / \underline{27.9} / 11.2 \\
AGuide (Ours) & \underline{97.2} / \textbf{97.1} / \underline{72.3} & \textbf{81.8} / \textbf{81.4} / \textbf{65.8} & \textbf{50.6} / \textbf{50.3} / \underline{29.1} & \textbf{35.1} / \textbf{34.2} / \underline{16.6} & \textbf{34.8} / \textbf{32.0} / \textbf{18.9} \\
\bottomrule
\end{tabular}

\par\vspace{4pt}

\begin{tabular}{@{}lccccc@{}}
\toprule
Method & 50--60\,m & 60--70\,m & 70--80\,m & 80--90\,m & 90--100\,m \\
GT count & 592 & 345 & 250 & 164 & 100 \\
\midrule
No Fusion (Veh.) & 16.4 / 14.8 / 9.4 & 23.4 / 21.5 / 11.3 & 20.5 / 19.5 / 11.1 & 27.3 / 26.1 / 14.0 & \underline{27.8} / \underline{26.8} / 8.8 \\
No Fusion (UAV) & 6.1 / 4.9 / 1.9 & 3.2 / 3.2 / 1.9 & 1.9 / 0.7 / 0.4 & 2.0 / 1.1 / 0.0 & 0.0 / 0.0 / 0.0 \\
Late Fusion & 19.6 / 18.5 / 9.9 & 19.9 / 19.9 / 12.5 & 15.5 / 14.6 / 8.4 & 19.1 / 19.1 / 4.7 & 10.1 / 10.1 / 7.1 \\
\midrule
CoBEVT & 23.9 / 18.4 / 7.7 & 21.8 / 21.6 / 12.6 & 20.7 / 19.8 / 13.1 & 33.0 / 31.4 / 15.1 & 20.9 / 20.0 / \underline{15.4} \\
Where2comm & 28.9 / 21.6 / 10.1 & 28.7 / 27.9 / 14.3 & 25.8 / 25.1 / \underline{13.7} & 32.9 / 30.5 / 18.9 & 21.3 / 19.4 / 12.2 \\
V2X-ViT & 21.2 / 16.9 / 7.1 & 23.6 / 21.1 / 11.5 & 21.9 / 19.3 / 10.8 & 24.3 / 23.2 / 12.7 & 1.3 / 1.3 / 0.0 \\
CoAlign & 32.2 / 28.5 / \underline{11.9} & \underline{32.0} / \textbf{31.0} / \textbf{19.0} & 26.8 / \underline{26.4} / 13.0 & \underline{38.0} / \textbf{38.0} / \textbf{23.4} & 26.7 / 26.1 / \textbf{17.3} \\
HEAL & 11.2 / 11.2 / 5.9 & 17.3 / 15.8 / 7.4 & 17.9 / 17.2 / 8.3 & 21.3 / 21.3 / 12.5 & 21.6 / 20.2 / 7.3 \\
AVCPNet$^\ddagger$ & \underline{35.7} / \underline{32.9} / 11.3 & 30.8 / \underline{29.9} / 8.4 & \underline{27.4} / 25.6 / 4.1 & 24.6 / 23.3 / 5.8 & 16.9 / 16.9 / 8.1 \\
AGuide (Ours) & \textbf{38.5} / \textbf{33.2} / \textbf{15.8} & \textbf{32.6} / 29.8 / \underline{16.8} & \textbf{29.9} / \textbf{29.1} / \textbf{18.5} & \textbf{39.9} / \underline{36.4} / \underline{21.2} & \textbf{29.0} / \textbf{29.0} / 15.3 \\
\bottomrule
\end{tabular}
\end{table*}

\section{Relation to Conditional Displacement Regression.}
Because AGuide removes the velocity networks at inference without performing ODE integration, its auxiliary objective functions as training-only representation regularization rather than generative transport. Under our paired linear path $z_t = (1-t)z_u + t z_v$, traditional conditional displacement regression (CDR) corresponds to the boundary special case at $t=0$, where the network predicts $z_v - z_u$ from $z_u$ alone (\ie, Fixed $t=0$ in Table~5 of the main paper). Sampling $t \sim \mathcal{U}(0,1)$ differs from single-endpoint CDR in both supervision domain and gradient dynamics. With frozen vehicle references $z_v$, the gradient of the Flow Matching loss with respect to the adapted UAV feature $z_u$ takes the form
\begin{equation}
\nabla_{z_u} \|v_\theta(z_t, t, c_p) - (z_v - z_u)\|_2^2
= 2 \bigl((1-t) J_{v_\theta}(z_t) + I\bigr)^\top e_t,
\end{equation}
where $J_{v_\theta}(z_t) = \partial v_\theta / \partial z_t$ and $e_t = v_\theta(z_t, t, c_p) - (z_v - z_u)$. Whereas CDR ($t=0$) applies unattenuated Jacobian feedback exclusively at the source endpoint $z_u$, continuous interpolation exposes $v_\theta$ to mixed cross-platform states across $t \in (0,1)$ (where directional alignment peaks in Table~\ref{tab:velocity_alignment}) while modulating the Jacobian term by $(1-t)$. This multi-state path supervision enforces consistent displacement directions along the interpolation trajectory and acts as a smoother regularizer on $z_u$, yielding higher strict-IoU AP than the $t=0$ boundary case.

\section{AVCPNet Reimplementation}
\label{app:baseline_details}

Since the authors have not released the official implementation, we reimplement AVCPNet~\cite{10909254} from its published description. The original method is camera-based. It uses Collaborative Depth Optimization (CDO) to refine image depth before BEV construction and Cross-Domain Cross-Adaptation (CDCA) to align and fuse vehicle and UAV BEV features.

\subsection{Camera Reimplementation on OPV2V-Air}

The original camera pipeline builds on BEVDet4D with a ResNet-50 backbone, Lift--Splat--Shoot view transformation, and a CenterPoint detection head. To compare both methods within the same camera detection framework on OPV2V-Air, we retain Lift--Splat--Shoot and use the same EfficientNet image encoder and anchor-based detection head as AGuide. We preserve the role of CDO as a semantic-guided depth refinement step before BEV lifting. Specifically, we implement a CRF-style update that uses feature affinities within each image to weight neighboring depth distributions. The refined depth distributions are used to lift image features into BEV, where CDCA performs air--ground feature fusion.

The original CDCA uses a four-level feature pyramid, correlation-based scale weighting, and BEV attention to combine the two views. Our implementation retains the four scales, uses cosine similarity with softmax normalization for scale weighting, and spatially pools the attention keys and values to reduce computation. 

\subsection{LiDAR Adaptation on V2U4Real}

For LiDAR inputs, we transform the point clouds into the ego vehicle frame before voxelization and use a shared PointPillar encoder to construct BEV features. This removes the need for image lifting and CDO. We fuse the resulting features using the same four-level CDCA design with pooled attention and predict boxes with an anchor-based detection head.

\section{Additional Qualitative Results}

\subsection{OPV2V-Air}
\label{app:qualitative_opv2v_air}

\begin{figure*}[t]
    \centering
    \begin{subfigure}[t]{0.49\textwidth}
        \centering
        \includegraphics[width=\linewidth]{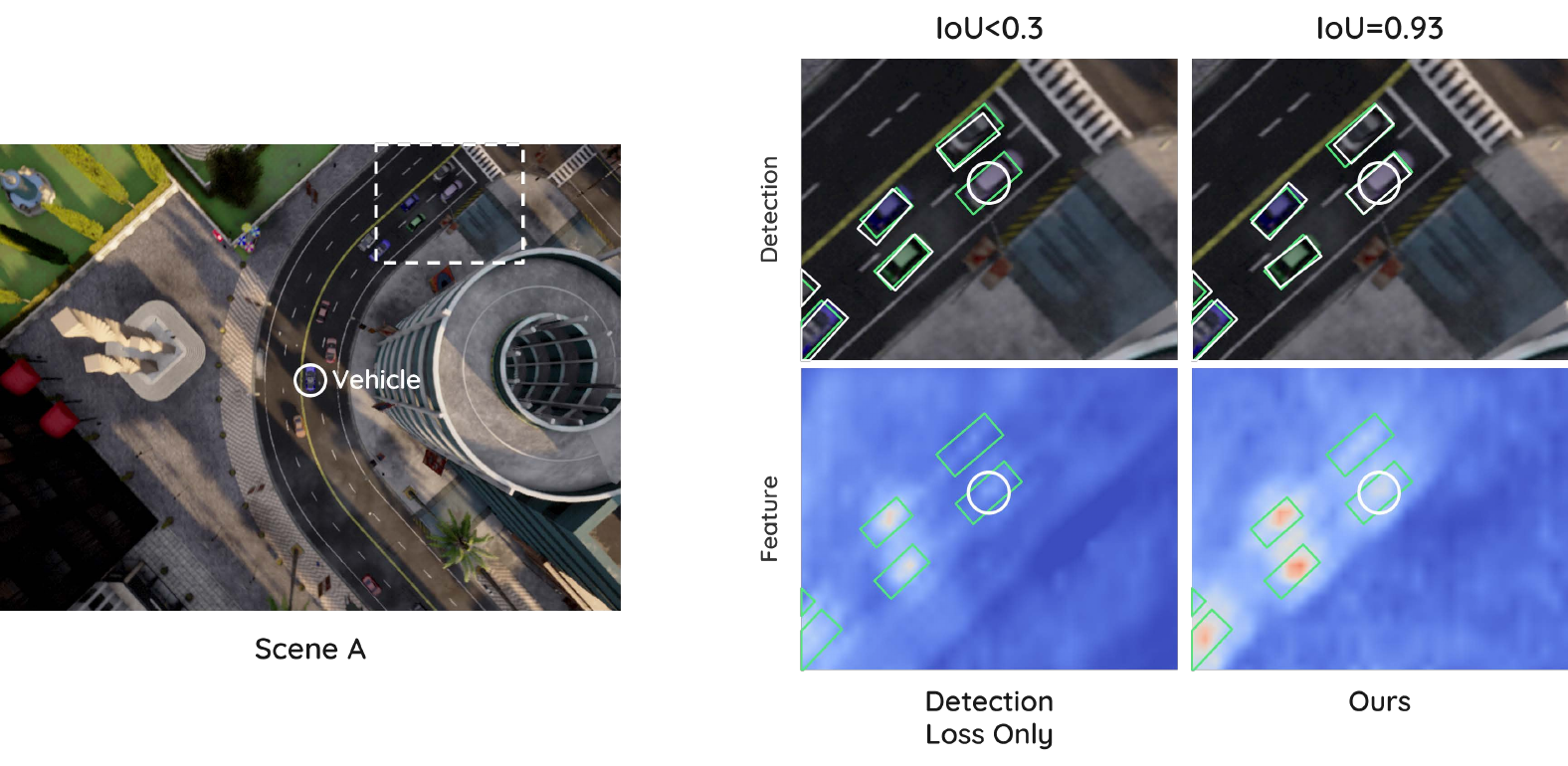}
        \caption{Long-range recovery on a curved urban road.
        AGuide detects a distant vehicle missed by the detection-only baseline.}
        \label{fig:opv2v_air_a}
    \end{subfigure}
    \hfill
    \begin{subfigure}[t]{0.49\textwidth}
        \centering
        \includegraphics[width=\linewidth]{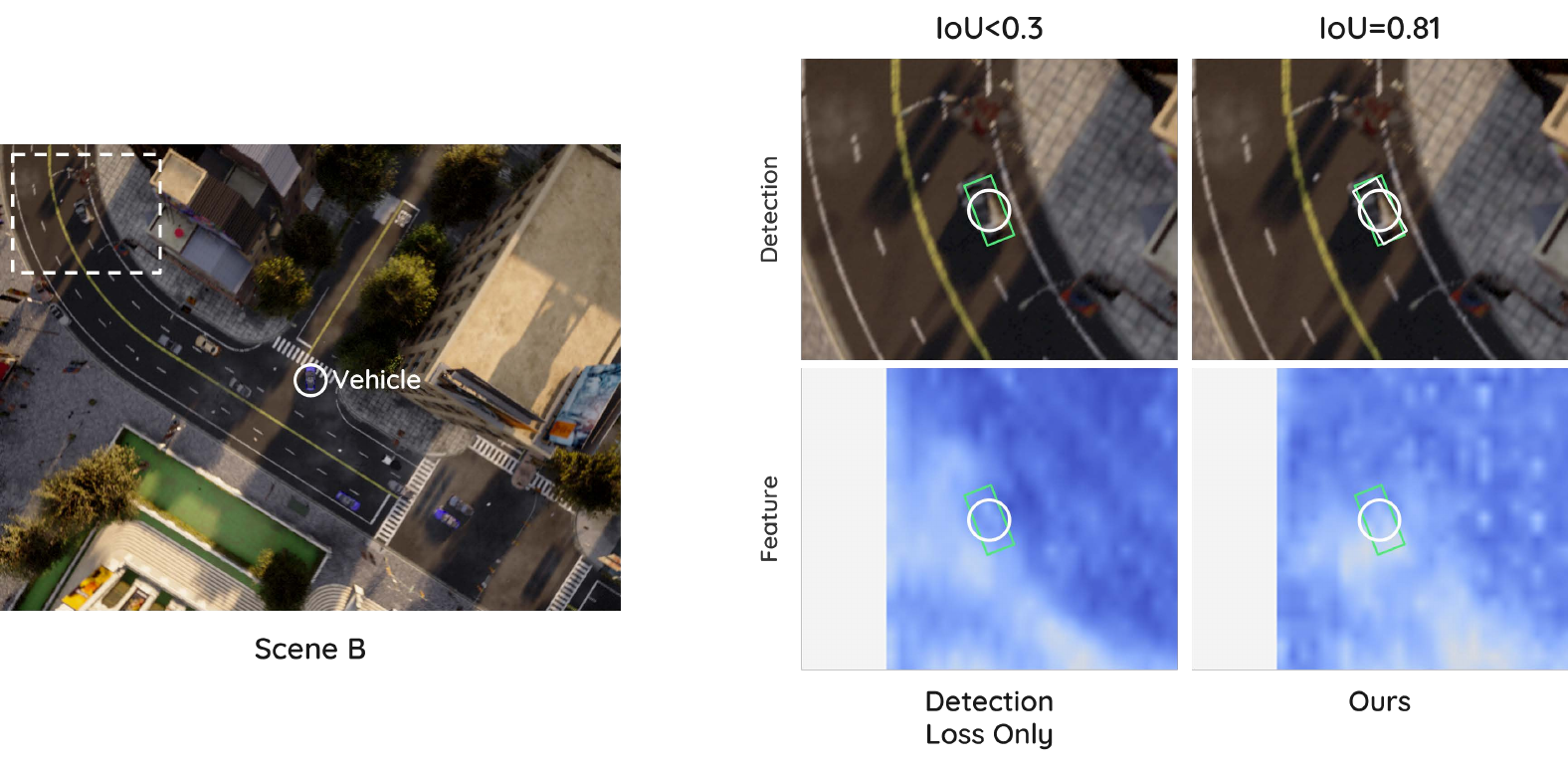}
        \caption{Long-range recovery at an urban intersection.
        AGuide recovers the highlighted vehicle across the intersection.}
        \label{fig:opv2v_air_b}
    \end{subfigure}

    \par\medskip

    \begin{subfigure}[t]{0.49\textwidth}
        \centering
        \includegraphics[width=\linewidth]{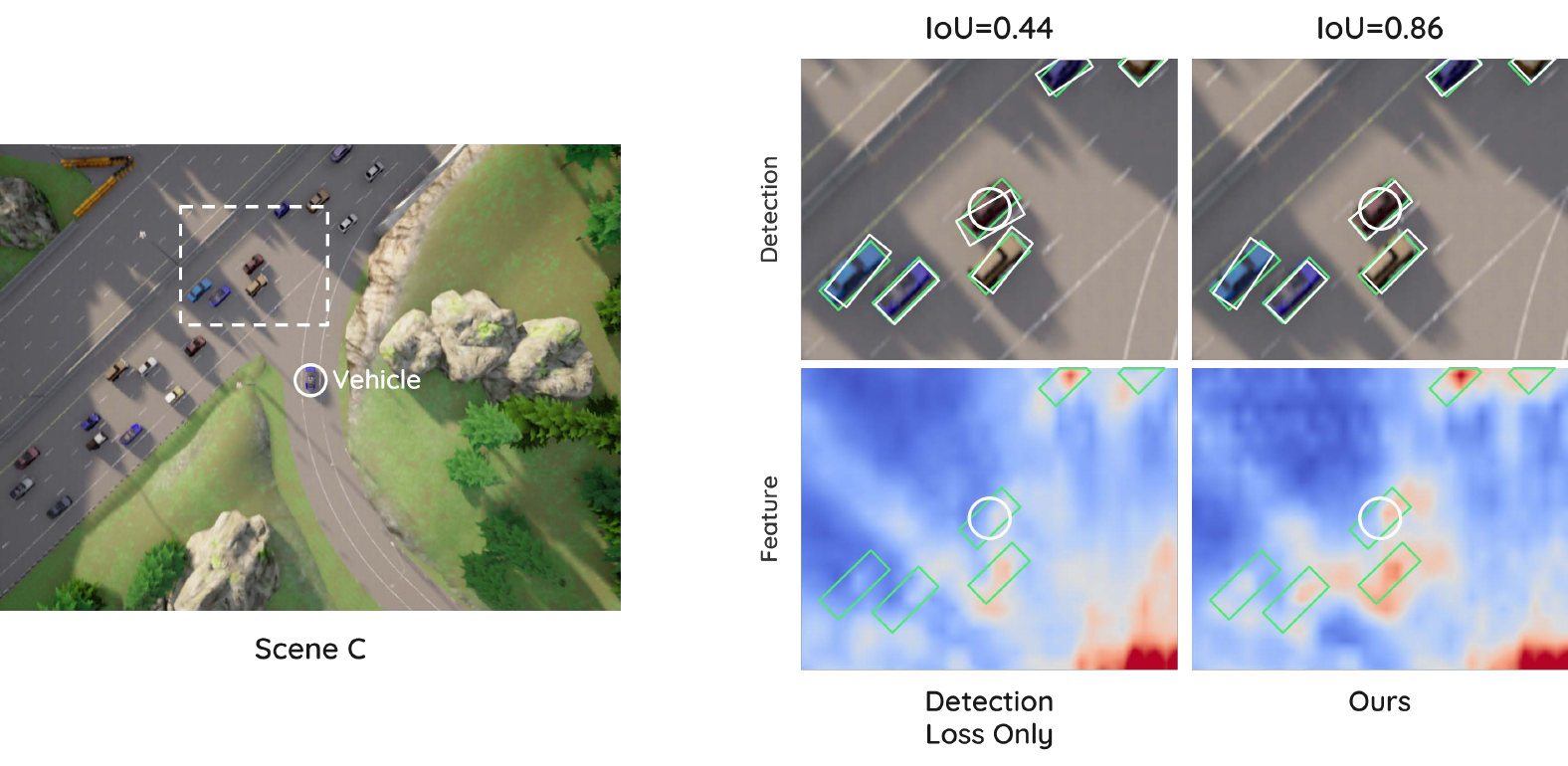}
        \caption{Near- to mid-range refinement on a curved road.
        AGuide improves the predicted box's fit to the highlighted vehicle.}
        \label{fig:opv2v_air_c}
    \end{subfigure}
    \hfill
    \begin{subfigure}[t]{0.49\textwidth}
        \centering
        \includegraphics[width=\linewidth]{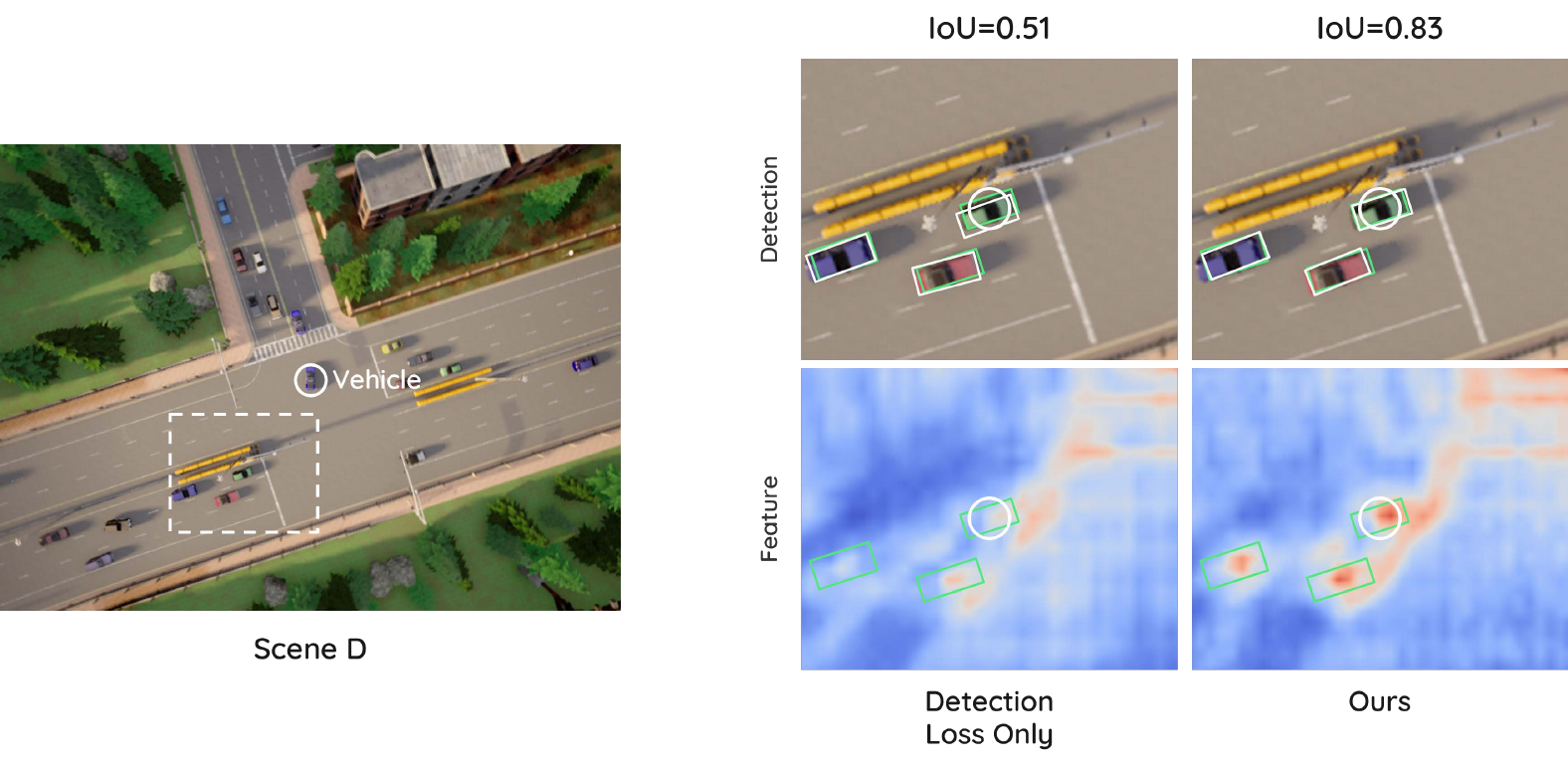}
        \caption{Near- to mid-range refinement on a multilane road.
        AGuide improves the overlap between the predicted and ground-truth boxes.}
        \label{fig:opv2v_air_d}
    \end{subfigure}

    \caption{Additional qualitative comparisons between detection-only
    supervision and AGuide on OPV2V-Air. Each panel shows an aerial
    overview, detection results, and BEV feature maps. The IoU values
    refer to the circled vehicles.}
    \label{fig:opv2v-air}
\end{figure*}

Figure~\ref{fig:opv2v-air} compares AGuide with detection-only supervision across four OPV2V-Air scenes. In (a) and (b), AGuide recovers distant vehicles that the baseline fails to match at BEV IoU~0.3. In (c) and (d), both methods detect the highlighted vehicles, and AGuide produces boxes that more closely overlap the ground truth. The fused BEV feature maps also show more distinct responses around the highlighted targets. These examples illustrate long-range recovery and improved localization of already detected vehicles.

\subsection{V2U4Real}
\label{app:qualitative_v2u4real}

\begin{figure*}[t]
    \centering
    \includegraphics[width=1\linewidth]{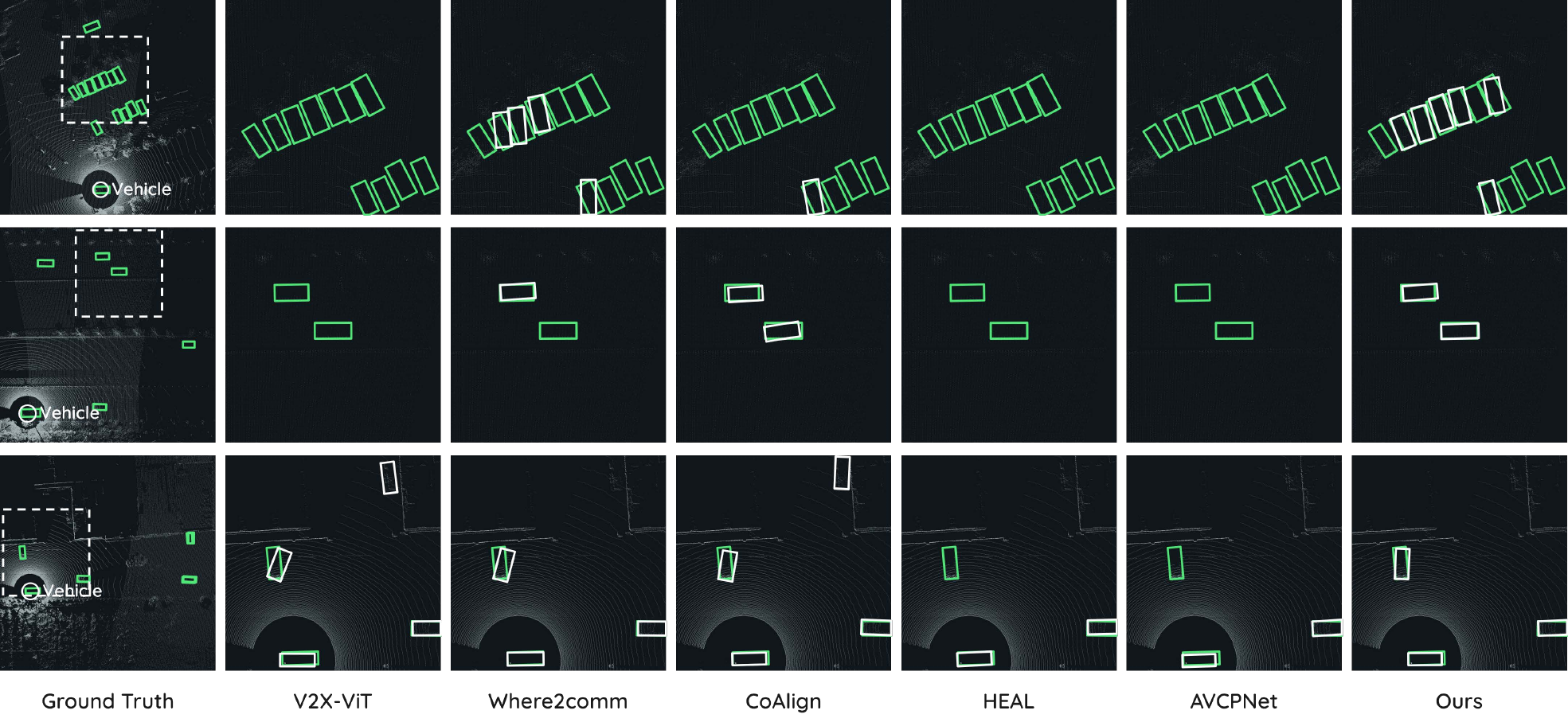}
    \caption{Qualitative comparison on V2U4Real with one vehicle and one UAV. Each row shows a different LiDAR scene. Dashed rectangles in the left column mark the regions enlarged in the remaining panels. Green boxes denote ground truth, and white boxes denote predictions. In these examples, AGuide detects vehicles missed by several baselines and produces boxes that more closely match the ground truth.}
    \label{fig:v2u4real}
\end{figure*}

Figure~\ref{fig:v2u4real} compares AGuide with other collaborative methods across three real-world LiDAR scenes. In the first two scenes, AGuide detects vehicles missed by several baselines, including targets in a closely spaced row. In the third scene, its predictions more closely overlap the ground-truth boxes, and it avoids the unmatched predictions seen in some baseline outputs. These examples illustrate how the detection and localization benefits extend to real-world LiDAR collaboration.

\end{document}